\documentclass{article}

\usepackage{hyperref}
\usepackage{authblk}
\usepackage{times}
\usepackage{booktabs}
\usepackage{subfigure}
\usepackage{amsfonts}       
\usepackage{microtype}   
\usepackage{graphicx}
\usepackage{amsmath}
\usepackage{float}
\usepackage{algorithm}
\usepackage[noend]{algpseudocode}
\usepackage[verbose=true,a4paper]{geometry}

\AtBeginDocument{
	\newgeometry{
		textheight=9in,
		textwidth=6in,
		top=1in,
		headheight=14pt,
		headsep=25pt,
		footskip=30pt
	}
}

\newcommand{\tabincell}[2]{\begin{tabular}{@{}#1@{}}#2\end{tabular}}

\title{CatVRNN: Generating Category Texts via Multi-task Learning}
\author{Pengsen Cheng \thanks{cps11@163.com}}
\author{Jinqiao Dai \thanks{djqqiao@hotmail.com}}
\author{Jiayong Liu \thanks{Corresponding Author: ljy@scu.edu.cn}}
\affil{Sichuan University}
\date{}

\begin{document}

\maketitle
\begin{abstract}
Controlling the model to generate texts of different categories is a challenging task that is receiving increasing attention. Recently, generative adversarial networks (GANs) have shown promising results for category text generation. However, the texts generated by GANs usually suffer from problems of mode collapse and training instability. To avoid the above problems, in this study, inspired by multi-task learning, a novel model called category-aware variational recurrent neural network (CatVRNN) is proposed. In this model, generation and classification tasks are trained simultaneously to generate texts of different categories. The use of multi-task learning can improve the quality of the generated texts, when the classification task is appropriate. In addition, a function is proposed to initialize the hidden state of the CatVRNN to force the model to generate texts of a specific category. Experimental results on three datasets demonstrate that the model can outperform state-of-the-art text generation methods based on GAN in terms of diversity of generated texts.
\end{abstract}

\section{Introduction}
Category text generation is a typical subfield of text generation which uses supernumerary information to generate coherent and meaningful text in different categories\cite{Liu2020}. As the next level expression of machine intelligence, the generated texts are more friendly to humans. With the rapid development of online interactive entertainment, category text generation has a wide range of applications, such as, virtual idols\footnote{See the link for details of related news. https://news.cgtn.com/news/2021-10-29/China-s-virtual-idols-worth-big-money-in-metaverse-era-14Kr8PeLRja/index.html}. Category text generation is an extension of emotional text generation and a type of conditional text generation\cite{Guo2021}, where the efficiently integration of additional category information with traditional model structures is a significant challenge\cite{Guo2021}. 

A few works employ generative adversarial networks (GANs) to generate category text. In GANs, generators are guided by a discriminator, which not only discriminates the authenticity of the generated texts but also discriminates the categories of the texts. The core target of the category text generation model is to learn the conditional distribution between words and categories. In fact, the latest GANs approximate the target using two-step training\cite{Liu2020,Wang2018,Chen2020}. In pre-training, the generators learn the relationship between words, that is, how to generate basic texts. In adversarial training, the generators learn the relationship between words and categories, that is, how to generate category texts. However, when humans write articles, they also learn to read. To learn new tasks, the knowledge acquired in learning related tasks is often applied.

Multi-task learning(MTL)\cite{Ruder2017} is a good solution for simulating this process. The primary task of MTL is to learn the relationship between words. The auxiliary task of MTL is to learn the relationship between words and categories. GANs have shown some positive results, but mode collapse and training instability are fundamental problems of GANs\cite{Liu2016, Kodali2017}. To solve the discrete nature of texts and pass the gradient from the discriminator to the generator, reinforcement learning (RL) and Monte Carlo (MC) searches have been adopted\cite{Yu2017,Guo2017}. Although these methods increase the training difficulty of GANs. In addition, to improve the diversity of the texts generated by GANs, these studies\cite{Liu2020, Chen2020} employ sample strategies. However, an improvement in diversity easily leads to a significant degradation of quality. 

In this study, the aim is to generate a variety of high-quality category texts using MTL. To avoid these issues, a new category text generation model called category-aware variational recurrent neural network (CatVRNN) is proposed. First, the variational recurrent neural network (VRNN) is divided into three parts: shared layers, generation task layers and classification task layers. By incorporating latent random codes into a recurrent neural network (RNN), VRNN addresses sequence modeling problems\cite{Chung2015}. The generation layers aim to generate samples that are as realistic as the real samples, while the classification layers aim to discriminate categories of real samples. The shared layers learn and adjust the knowledge by back propagation from the generation and classification tasks, simultaneously. Second, a novel function is proposed with hyperparameters to initialize the hidden state of CatVRNN. It is important to calculate a ``good" initialization hidden state for CatVRNN. The initialization hidden state needs to reflect the gap between the categories, which activate the neurons to learn different distributions\cite{Mohajerin2017}. However, this gap must be kept within a certain range to preclude the classification layers from discriminating texts based only on the hidden state. Third, an adaptive hidden state initialization method is proposed to solve the problem of hyperparameters. Finally, the impact of the classification task on the generation task in MTL is discussed, indicating that the quality of the generated texts can be improved by an appropriate classification task.

In summary, the contributions are:
\begin{itemize}
	\item[1)] 
	A novel method for generating category texts is proposed using MTL. To the best of our knowledge, this is the first study to generate category texts by training a model as a generator and classifier simultaneously. 
	\item[2)] 
	A new function is proposed to initialize the hidden state of the RNN, which can activate different neurons to generate different categories of texts.
	\item[3)] 
	Extensive experiments are performed on three datasets, whereby the results demonstrate the efficacy and superiority of the proposed model.
\end{itemize}

The rest of the paper is organized as follows: In Section 2, the literature on both models and on tasks related to the research work are provided. In Section 3, CatVRNN is described. In Section 4, the effectiveness of the proposed model is validated. In Section 5, the impact of the classification on the generation task is discussed. Concluding remarks are presented in Section 6.

The following table contains a glossary of the mathematical symbols used in the paper.

\begin{table}[H]
	\centering
	\caption{The definitions of symbols.}
	\label{tab.symbols}
	\resizebox{0.6\textwidth}{!}{
		\begin{tabular}{|l|l|}
			\hline
			Symbol & Definition\\
			\hline
			$K$ & number of categories in the dataset \\
			$x$ & a word in the input \\
			$X$ & a sentence of input \\
			$S$ & the length of $X$ \\
			$y$ & a word in the output \\
			$Y$ & a sentence of output \\
			$T$ & the length of $Y$ \\ 
			$C$ & the category of the sentence \\
			$\nu$ & the vocabulary of candidate words\\
			$z$ & the latent random codes of VAE \\
			$\mu$ & mean of the distribution \\
			$\sigma$ & standard deviation of the distribution \\
			$diag$ & a square diagonal matrix \\
			$\mathcal N$ & normal distribution \\
			$\varphi_{\tau}^{prior}$ & conditional prior \\
			$\varphi_{\tau}^{enc}$ & approximate posterior \\
			$\varphi_{\tau}^{dec}$ & generating distribution \\
			$\varphi_{\tau}^{z}$ & feature extractor of $z$ \\
			$\varphi_{\tau}^{x}$ & feature extractor of $x$ \\
			$h$ & the hidden state of VRNN\\
			$f_{\theta}$ & recurrence function of RNN with parameter $\theta$ \\
			$KL$ & Kullback–Leibler divergence\\			
			$\phi$ & the initialization function of the hidden state \\		
			$\vec{r}$ & a vector sampled from uniform distribution \\
			$\omega$ & slope of $\phi$\\
			$b$ & intercept of $\phi$\\
			$U$ & uniform distribution\\				
			\hline
		\end{tabular}
	}
\end{table}

\section{Related Work}
GANs are widely employed in text generation, because traditional RNN-based text generation models always suffer from the exposure bias problem\cite{Bengio2015}. Generators can be guided to simulate real data distribution by adversarial training. Thus, adversarial ideas can be encountered in most categories of text generation. GANs are designed to process differentiable data rather than discrete data; however, text is an example of typical discrete data. To solve this problem, SeqGAN and LeakGAN apply RL as an update strategy for the generator\cite{Yu2017, Guo2017}. To control generation, CSGAN concatenates the controllable information and token distribution to obtain the prior information\cite{Li2018}. However, these models lack diversity in the generated texts because of mode collapse. SentiGAN addresses this problem by training multiple generators simultaneously and creates a new penalty\cite{Wang2018}. In SentiGAN, each generator generates the text of a specific sentiment label. However, as the number of categories increase, the parameters to be trained are more numerous. To decrease training complexity, CatGAN evolves a population of generators that combine various mutation strategies in an environment using a hierarchical evolutionary learning algorithm\cite{Liu2020}. In CTGAN, an emotion label is adopted as an input channel to specify the output text, and an automated word-level replacement strategy is adopted to guarantee the quality and diversity of the text\cite{Chen2020}. Variational autoencoder (VAE) based methods also adopt the adversarial idea to generate category texts. Hu et al.\cite{Hu2017} combined VAE and holistic attribute discriminators for the effective imposition of semantic structures, similar to CSGAN,  whereby the outputs of the discriminators are concatenated as controllable information with the token distribution as the prior.

Few researchers have already attempted to apply MTL to solve related problems. Lu et al.\cite{Lu2019} proposed an abstractive text summarization model that extends regularization using MTL to perform additional text categorization and syntax annotation tasks. Nishino et al.\cite{Nishino2020} introduced an MTL model with a shared encoder and multiple decoders to generate headlines, key phrases, summaries, and category outputs consistently. Zhu et al.\cite{Zhu2020} trained nurture language generation and language modeling tasks simultaneously to generate high-quality responses in task-oriented dialogues. Xu et al.\cite{Xu2020} proposed a key information guide network for abstractive text summarization based on an MTL framework. The primary task of this model is abstraction based on the encoder-decoder structure, and the second task is key sentence extraction. 

For text generation tasks, GAN is an excellent option. Therefore, almost all category text generation models are currently based on GANs. However, GAN has several inherent challenges that are difficult to solve, such as the complexity of the training process and model collapse. Among these, the problem of model collapse has a direct impact on the diversity of generated texts. Diversity of expression is a type of model creativity requiring different ways of thinking to solve the problem of category text generation. CatVRNN proposed in this study extends VRNN with MTL to solve the category text generation problem, whereby an initialization function of the hidden state is proposed to force generation, and the feedback from the classification corrects the generation.  

\section{Proposed Method}
The category text generation task is denoted as follows: Given a dataset with $K$ categories, suppose the goal is to generate a sentence  $Y=(y_{1},y_{2},...,y_{T})$ from the input sentence $X=(x_{1},x_{2},...,x_{S})$ in the specific category $C$, with $y, x\in\nu$. $Y$ is sampled by Eq.(\ref{eq.define}).

\begin{equation}
	p(Y|X,C)=\prod_{t=1}^{T}p(y_{t}|X,C,y_{<t}) 
	\label{eq.define}
\end{equation}

The mathematical description of category text generation using MTL is as follows: Although $p(x|C)$ is extremely important, it is difficult to calculate $p(x|C)$ directly using a single generation task. However, it is easy for the classification task to calculate $p(C|x)$. Then, $p(x|C)$ can be calculated from $p(C|x)$ using Bayes' theorem. The generation task can thus accurately sample words with category features using $p(x|C)$. 

\subsection{Variational Recurrent Neural Network}
The variational recurrent neural network is extended to generate sequences using an encoder and a decoder within the VAE framework\cite{Chung2015}. Generating a sample using only a traditional RNN involves nondeterministic operations in the output space. VRNN also has random variability at a potentially more abstract level as captured by the VAE latent codes. This has been applied in some generation tasks, such as conditional text generation\cite{Xu2018}, graph generation\cite{Su2019}, and dialogue generation\cite{Qiu2020}. 

Each operation of VRNN can be decomposed as follows.
\begin{itemize}
	\item[1)] 
	\textbf{Prior}: The conditional prior is computed using Eq.({\ref{eq.vrnn_prior}}).
	\begin{equation}
		z_{t} \sim \mathcal N(\mu_{0,t},diag(\sigma_{0,t}^{2})),\text{ where }[\mu_{0,t},\sigma_{0,t}]=\varphi_{\tau}^{prior}(h_{t-1})
		\label{eq.vrnn_prior}
	\end{equation} 
	\item[2)]
	\textbf{Generation}: The generating function uses Eq.(\ref{eq.vrnn_generating}), and the generative model results in factorization, as shown in Eq.(\ref{eq.vrnn_generation_factorization}).
	\begin{equation}
		x_{t}|z_{t} \sim \mathcal N(\mu_{x,t},diag(\sigma_{x,t}^{2})),\text{ where }[\mu_{x,t},\sigma_{x,t}]=\varphi_{\tau}^{dec}(\varphi_{\tau}^{z}(z_{t}), h_{t-1})
		\label{eq.vrnn_generating}
	\end{equation}
	\begin{equation}
		p(x\le T, z\le T)=\prod_{t=1}^{T}[p(x_{t}|z_{\le t},x_{<t})\cdot p(z_{t}|x_{<t},z_{<t})]
		\label{eq.vrnn_generation_factorization}
	\end{equation}
	\item[3)] 
	\textbf{Recurrence}: To update the RNN hidden state uses Eq.(\ref{eq.vrnn_rnn}), where $f$ is a deterministic nonlinear transition function and $\theta$ is the parameter set of $f$.
	\begin{equation}
		h_{t}=f_{\theta}(\varphi_{\tau}^{x}(x_{t}),\varphi_{\tau}^{z}(z_{t}),h_{t-1})
		\label{eq.vrnn_rnn}
	\end{equation} 
	\item[4)]
	\textbf{Inference}: To infer the approximate posterior, Eq.(\ref{eq.vrnn_inference}) is used along with the posterior results of factorization, as shown in Eq.(\ref{eq.vrnn_inference_factorization}).
	\begin{equation}
		z_{t}|x_{t}\sim \mathcal N(\mu_{z,t},diag(\sigma_{z,t}^{2})),\text{ where }[\mu_{z,t},\sigma_{z,t}]=\varphi_{\tau}^{enc}(\varphi_{\tau}^{x}(x_{t}),h_{t-1})
		\label{eq.vrnn_inference}
	\end{equation}
	\begin{equation}
		q(z_{\le T}|x_{\le T})=\prod_{t=1}^{T}q(z_{t}|x_{\le t},z_{<t})
		\label{eq.vrnn_inference_factorization}
	\end{equation} 
\end{itemize}

In Eq.(\ref{eq.vrnn_rnn}), $h_t$ is a function of $x \le t$ and $z \le t$. Therefore, Eq.(\ref{eq.vrnn_generating}) and Eq.(\ref{eq.vrnn_inference}) define the $p(x_{t}|z \le t,x<t)$ and $p(z_{t}|x<t,z<t)$ distributions, respectively; $\varphi_{\tau}^{enc}$ and  $\varphi_{\tau}^{dec}$ can be highly flexible functions, such as neural networks; $\varphi_{\tau}^{x}$ and $\varphi_{\tau}^{z}$ can also be neural networks that extract features from $x_{t}$ and $z_{t}$. The learning objective function is a time-step-wise variational lower bound calculated using Eq.(\ref{eq.vrnn_generating}) and Eq.(\ref{eq.vrnn_inference}).

\begin{equation}
	\mathbb{E}_{q(z_{\le T}|x_{\le T})}[\sum_{t=1}^{T}(-KL(q(z_{t}|x_{\le t},z_{<t})||p(z_{t}|x_{\le t},z_{<t}))+logp(y_{t}|z_{\le t},x_{<t}))] 
	\label{eq.vrnn_learning}
\end{equation} 

\subsection{Adjustment of VRNN}
The variational recurrent neural network, which is a type of generative model, is proposed to model the joint probability distribution of sequences. For VRNN to model the conditional probability distribution between sequences and additional category information, it must be adjusted to respond to new situations. The hidden state of VRNN can be used to build a classifier on its structure, to turn VRNN into an MTL model.  

\begin{figure}[H]
	\centering
	\subfigure[adjusted structure of VRNN]{
		\includegraphics[width=2in]{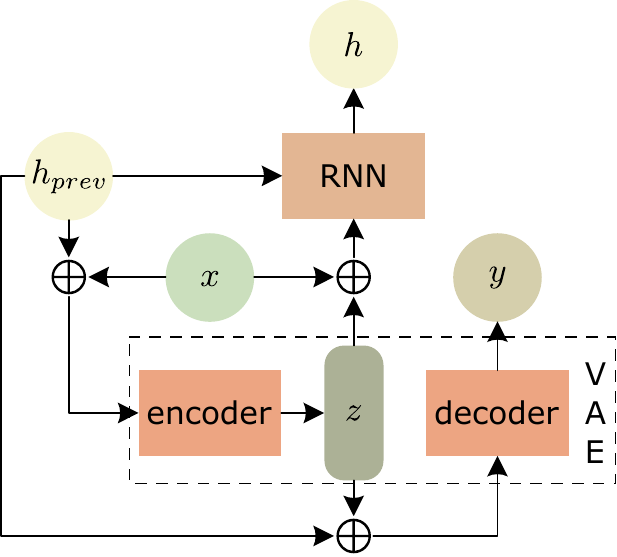}
		\label{fig.vrnn_cell}
	}
	\subfigure[structure of VAE in VRNN]{
		\includegraphics[width=2in]{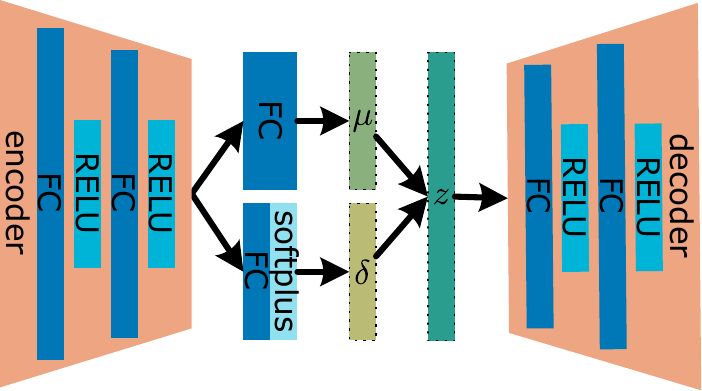}
		\label{fig.vae}
	}
	\caption{The core structure of CatVRNN. $\oplus$ represents the concatenation of vectors. The VAE in the (b) figure is depicted in the (a) figure. FC denotes fully connected layer. $\mu$ and $\sigma$, which are got by transforming the output of encoder, denote the mean and standard deviation of $z$, respectively.}
	\label{fig.catvrnn_cell}
\end{figure}

The core structure of CatVRNN, which is the adjusted structure of VRNN, is shown in Fig.\ref{fig.vrnn_cell}. The most important adjustment is that the conditional prior is no longer computed using Eq.(\ref{eq.vrnn_prior}), which means that the posterior $p(z|x)$ is not forced to be as close to the prior $p(z)$ as possible. In fact, the KL term is the cause, while the conditional prior is the result. In Eq.(\ref{eq.vrnn_learning}) the KL term is calculated using the conditional prior and approximate posterior. When the KL term dose not need to be calculated, the conditional prior dose not need to be calculated, either. The specific reasons are explained in the learning objective subsection.

The feature extractors of $x_{t}$ and $z_{t}$ are also cancelled. In text sequences, $x_{t}$ is the embedding of a single word or token that does not contain complex features. The nonlinear transition of $\varphi_{\tau}^{x}$ and $\varphi_{\tau}^{z}$ can destroy the information in $x_{t}$ and $z_{t}$. Therefore, the adjusted VRNN updates the hidden state using Eq.(\ref{eq.catvrnn_h}) and approximates the posterior using Eq.(\ref{eq.catvrnn_z}).

\begin{equation}
	h_{t}=f_{\theta}(x_{t} \oplus z_{t},h_{t-1})
	\label{eq.catvrnn_h}
\end{equation} 

\begin{equation}
	z_{t}|x_{t}\sim\mathcal N(\mu_{z,t},diag(\sigma_{z,t}^{2})),\text{ where }[\mu_{z,t},\sigma_{z,t}]=\varphi_{\tau}^{enc}(x_{t},h_{t-1})
	\label{eq.catvrnn_z}
\end{equation}   

The computational flow of the single-step CatVRNN is shown in Algorithm \ref{alg.catvrnn_cell}. In this flow, the calculation of $\mu$ and $\sigma$ are simplified.

\begin{algorithm}[H]
	\caption{Computational flow of single step CatVRNN}
	\small
	\label{alg.catvrnn_cell}
	\begin{algorithmic}[1]
		\Require the hidden state of the last time step $h_{prev}$, the input of the current time step $x$ 
		\Ensure the hidden state of the current time step $h$, the output of the current time step $y$
		\Function{catvrnn\_cell}{$h_{prev}, x$}
		\State $z\leftarrow encoder(x\oplus h_{prev})$
		\State $y\leftarrow decoder(z\oplus h_{prev})$
		\State $h\leftarrow RNN(x\oplus z)$
		
		\State \Return $h,y$
		\EndFunction
		
	\end{algorithmic}
\end{algorithm}

\subsection{Learning Objective}

\begin{figure}[H]
	\centering
	\includegraphics{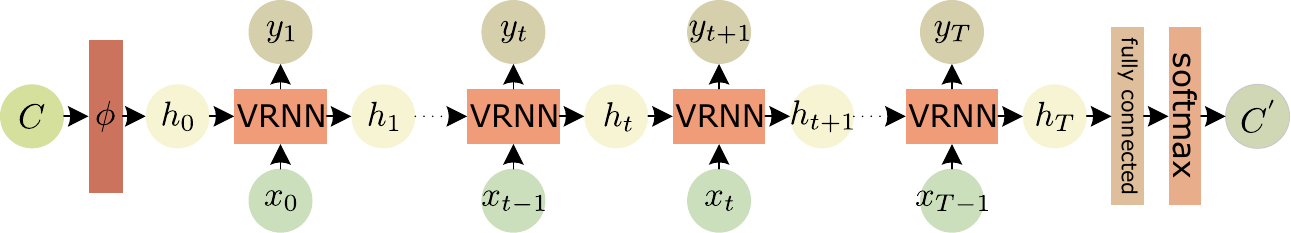} \\[\abovecaptionskip]
	\caption{The $T$-step unfolding structure of CatVRNN. $x_{0}$ is a padding token means starting, and if the length of $X$ is less than $T$, fill the length of $X$ to $T$ with padding token. The input is the filled $X=(x_0,...,x_{T-1})$ in the specific category $C$. The output contains the predicted sequence $Y=(y_1,...,y_{T})$ and the predicted category $C^{'}$. The classification task layers are constructed by full connected layer and softmax layer.}
	\label{fig.catvrnn}
\end{figure}

The $T$-step unfolding structure of CatVRNN is illustrated in Fig.\ref{fig.catvrnn} and the computational flow of the $T$-step is displayed in Algorithm \ref{alg.catvrnn}. CatVRNN comprises three core parts: the hidden state initialization function $\phi$, the adjusted VRNN, and the classification layers. The function $\phi$ initializes the hidden state $h_{0}$ by category $C$ of $X$. Each time step of VRNN outputs the hidden state $h$ and token $y$. 

\begin{algorithm}[H]
	\caption{$T$-step computational flow of CatVRNN}
	\small
	\label{alg.catvrnn}
	\begin{algorithmic}[1]
		\Require the sequence $X$, the category $C$, the length $T$ of the output sequence  
		\Ensure the output sequence $Y$, the predicted category $C^{'}$
		\Procedure{catvrnn}{$X,C,T$}
		
		\State Initialize a sequence $Y$ without item
		\State $h_{prev}\leftarrow \phi(C)$\Comment{initialize the hidden state}
		\If{the length of $X<T$}
		\State pad $X$ with padding token to length $T$	
		\EndIf
		
		\For{each $x\in X$}\Comment{generation task}
		\State $h_{prev},y\leftarrow CATVRNN\_CELL(h_{prev},x)$\Comment{Algorithm \ref{alg.catvrnn_cell}}
		\State pad $y$ to the end of $Y$
		\EndFor
		
		\State $C^{'}\leftarrow softmax(linear(h))$ \Comment{classification task}
		
		\State \Return $Y,C^{'}$
		\EndProcedure
		
	\end{algorithmic}
\end{algorithm}

The hidden state $h$ is the most important output of an RNN because it contains a large amount of information. In traditional RNNs, the final realization of all tasks is based on $h$. However, in VRNN, $h$ is just a bridge between the encoder and decoder. To take advantage of $h$ as much as possible, CatVRNN achieves text classification by $h$, because the classification task is an easy method in learning the distribution $p(C|x)$. Therefore, in the last time step of training, the hidden state $h_{T}$ is processed by classification layers, consisting of a fully connected layer and a softmax layer to calculate the probability of $C$. 

The classification loss for a given sequence of $X$ with category $C$ is then just the loss in the last time step, and $h_{T}$ is a function of $x \le T$ and $z \le T$. The learning objective is expressed with a negative log-likelihood, then:

\begin{equation}
	-logp_{classification}(C|x_{\le T},z_{\le T}) 
	\label{eq.calssification}
\end{equation}

The learning objective function of CatVRNN uses Eq.(\ref{eq.vrnn_learning}) and Eq.(\ref{eq.calssification}).
\begin{equation}
	\mathbb{E}_{q(z_{\le T}|x_{\le T})}[\sum_{t=1}^{T}logp_{generation}(x_{t}|z_{\le t},x_{<t})-logp_{classification}(C|z_{\le T},x_{\le T})] 
	\label{eq.loss}
\end{equation}

CatVRNN learns the generative model and classification model jointly by minimizing Eq.(\ref{eq.loss}) for the parameters. In Eq.(\ref{eq.vrnn_learning}), the KL term can be interpreted as a regularizer that prevents the inference network from copying $x$ into $z$, and in the case of a Gaussian, the prior and posterior have a closed-form solution\cite{Kingma2014}. However, the conditional prior is ignored and the KL divergence term is replaced with the learning objective of the classification task. The reasons for this are as follows:

First, the noise produced by the Gaussian prior can decrease the possibility of the posterior collapse of VAE, but the noise may cause non-convergence of classification tasks\cite{Minaee2021}. 

Second, if CatVRNN learns the generative, inference, and classification models, the classification task interferes with the posterior getting close to the Gaussian prior. However, He et al.\cite{He2019} found that during the initial stages of training, the inference network fails to approximate the true posterior of the model. As a result, the model is encouraged to ignore the latent encoding, and posterior collapse occurs. 

Finally, in CatVRNN, the elimination of the KL term can be seen as forcing the KL term to zero. The vanishing of the KL term may be a symptom of posterior collapse. One cause of the vanishing KL term is that in the standard VAE, the decoder capacity is too large. Because the decoder is too powerful, it may learn to ignore $z$ and instead rely solely on the autoregressive properties of $x$, causing $x$ and $z$ to be independent\cite{Goyal2017}. A lot of research is devoted to this problem, based on weakening the decoder\cite{Bowman,Bachman2016,Chen2016,Semeniuta2017}. To avoid $z$ being disregarded by the decoder and the decoder reconstructing the distribution exclusively by $x$, the decoder is weakened, causing the decoder to lose the ability to rebuild directly, being forced to adopt $z$.

As a result, CatVRNN does not consider the KL term as a learning objective by giving up a conditional prior and replacing it with the classification learning objective. A weakened decoder is adopted instead to reduce the impact of the missing KL term. 

The structure of the encoder and decoder are shown in Fig.\ref{fig.vae}. The VAE is mainly composed of an encoder and a decoder which are comprised of fully connected (FC), ReLU, FC, and ReLU layers, in that order. The role of each layer is the same for both encoder and decoder. The first FC layer implements linear transformations on the data to compress the data dimension; the second ReLU layer implements nonlinear transformations on the data to prevent the problem of vanishing gradient; and the third and final layers enhance the depth of the networks to better fit the data. The FC layer between the encoder and decoder calculates $\mu$ of $z$ using the encoder output; the FC and Softplus layers between the encoder and decoder are used to calculate $\sigma$ of $z$, using the encoder output.

\subsection{Initialization of Hidden State}
The CatVRNN was trained with a different initialization for each category. To control the model to generate category texts, the hidden state initialization was treated as a switch. To generate a certain category of text, the hidden state was initialized to a certain value. For example, assume that the initial values of the hidden state corresponding to the $C_{0}$ and $C_{1}$ category texts are $h^{0}_{0}$ and $h^{1}_{0}$, respectively. To generate the $C_{0}$ category text, $h^{0}_{0}$ is adopted to initialize the hidden state of the model, and to generate the $C_{1}$ category text, the hidden state of the same model is initialized to $h^{1}_{0}$.  

In RNN, the hidden state of each time step is related to the previous time step, as shown in Eq.(\ref{eq.catvrnn_h}). RNN realizes information accumulation and transmission, allowing the RNN recall of all prior information at any time step. This important theoretical foundation controls the RNN model, to generate category texts using the hidden state. When the initial state of the hidden state $h_{0}$ is initialized to $0$, the RNN focuses only on the sequence data and not on the external environment. However, through the sequence category, a distinct external environment is provided, and $h_{0}$ is initialized to record the category information. This is another foundation controlling the RNN model in terms of generating category texts by initializing the hidden state. Then, taking advantage of the structure of RNN, category information is transmitted from beginning to end. To train an RNN for steady-state response, a method is required to initialize the RNN states properly\cite{Mohajerin2017}.

\begin{equation}
	\phi(C,\vec{r},\omega)_{static}=\omega (-1)^{C} softmax(\vec{r})
	\label{eq.init_func}
\end{equation}

\begin{equation}
	softmax(r_{i})=\frac{e^{r_{i}}}{\sum_{j}^{|\vec{r}|}e^{r_{j}}}
	\label{eq.softmax}
\end{equation}

In Eq.(\ref{eq.init_func}), $\vec{r}$ is sampled from a uniform distribution, where $r_{i} \sim U[0,1)$. After normalizing $\vec{r}$ by $Softmax$, the hyperparameter $\omega$ is multiplied by $C$th power of -1. Essentially, Eq.(\ref{eq.init_func}) aims to maintain the hidden states of different categories at a constant difference. The noise generated by the uniform distribution is added to the initial hidden state for the regularization and stabilization of the dynamics of RNN\cite{Haykin2007}.

Eq.(\ref{eq.init_func}) is a simple way to reveal the influence of different initial hidden state. However, this method has a limitation. When $C$ is a natural integer, the $C$th power of $-1$ can only be $-1$ or $1$. Therefore, Eq.(\ref{eq.init_func}) has only two sets of valid values, when $\omega$ is set to a fixed value. This implies that CatVRNN with Eq.(\ref{eq.init_func}) can only generate two categories of text, because one set of valid values can only be the $h_0$ of one category. However, this model is not scalable. In addition, the value of hyperparameter $\omega$ determines the result, and determining the appropriate $\omega$ is difficult, although there might be an $\omega$ that can activate the best state. 

To this end, an adaptive method is proposed in this study, to initialize the hidden state, to expand $\phi$ without explicit hyperparameters to an arbitrary number of categories.

\begin{equation}
	\phi_{adaptive}(C; \omega ,b)= C \omega^{T}+b
	\label{eq.phi_ada}
\end{equation}

\begin{equation}
	\phi_{adaptive}(C; \omega ,b)=C \omega^{T}+b+ \vec{r}
	\label{eq.phi_ada_noise}
\end{equation}

The basic adaptive method is given by Eq.(\ref{eq.phi_ada}), where $\omega$ and $b$ are learned through back propagation. The goal of Eq.(\ref{eq.phi_ada}) is the same as that of Eq.(\ref{eq.init_func}), that is to maintain the hidden states of different categories at a constant difference. In training, in Eq.(\ref{eq.phi_ada_noise}), the noise term is $\vec{r}$, where $r_{i} \sim U[0,1)$, is added to regularize and stabilize the RNN. 

\section{Experiments}
\subsection{Experiment Setup}
The problem of generating long sentences is challenging in terms of text generation. Therefore, the focus in this work is on generating long sentences ($15 \le$length$\le 30$) from two categories on three real datasets.

\textbf{Movie reviews (MR)}\cite{Socher2013} have two sentiment classes (negative and positive). The original dataset consisted of 10,662 sentences. The sentences containing 15 and 30 words were randomly selected. The dataset thus generated contained 3,037 positive sentences and 3,048 negative sentences with a vocabulary size of 16,412.

\textbf{Beer reviews (BR)}\cite{McAuley2012} have two sentiment classes (negative and positive). The original dataset consisted of 1,180,821 sentences. The sentences containing 15 and 30 words were randomly selected. The dataset thus generated contained 6,000 positive sentences and 6,000 negative sentences with a vocabulary size of 12,844.

\textbf{Amazon reviews (AR)}\cite{McAuley2015} carries two types of products (books and applications). The original dataset consisted of 200,000 sentences. The sentences containing 15 and 30 words were randomly selected. The dataset thus generated contained 10,000 book sentences and 10,000 application sentences with a vocabulary size of 5037.

The models were trained on each dataset, and word embeddings were randomly initialized to a dimension of 300. RNNs were set as a single-layer GRU with a hidden dimension size of 256 and a maximum length of 30 words. VAEs were set to a latent code dimension size of 128. The optimization algorithm used was Adam. The models were trained for 250 epochs. We implemented our model based on Pytorch and the repeatable experiment code is made publicly available on https://github.com/cps11/catvrnn.

\textbf{Baselines} constitute the category text generative task of almost all existing work based on GAN. Therefore, using the following two baselines is adequate to demonstrate the advantages and disadvantages of GANs. In addition, CatVRNN is compared against the baselines, to objectively reflect the performance of CatVRNN.

\begin{itemize}
	\item[$\bullet$] 
	\textbf{SentiGAN}\cite{Wang2018} is a state-of-the-art sentimental text generative framework with multiple generators and a multi-class discriminator.
	\item[$\bullet$] 
	\textbf{CatGAN}\cite{Liu2020} is a relatively new category text generative model with a hierarchical evolutionary learning algorithm.
\end{itemize}

The generators of the GANs were pre-trained for 120 epochs and adversarial trained for 130 epochs as \cite{Yu2017,Wang2018}. A state-of-the-art text classifier\cite{Kim2014} was used to evaluate the category accuracy of the generated texts automatically. Each model was used to generate 5 K different categories of texts.  

Comparative experiments were conducted with several variants of the model to illustrate the effectiveness of the relevant improvements and innovations. The value of hyperparameter $\omega$ in Eq.(\ref{eq.init_func}) was set to 8.5, based on our experience. 

\begin{itemize}
	\item[$\bullet$] 
	\textbf{CatVRNN-nophi} is a model with a hidden state initialized to zero during training and to Eq.(\ref{eq.init_func}) during evaluation.	
	\item[$\bullet$] 
	\textbf{CatVRNN-static} is a model with a hidden state initialized using Eq.(\ref{eq.init_func}).	
	\item[$\bullet$] 
	\textbf{CatVRNN-adaptive} is a model with a hidden state initialized using Eq.(\ref{eq.phi_ada_noise}) during training and Eq.(\ref{eq.phi_ada}) during evaluation.
	\item[$\bullet$]
	\textbf{VRNN-static} is a model without classification layers, and the hidden state is initialized using Eq.(\ref{eq.init_func}).	  	
	\item[$\bullet$] 
	\textbf{VRNN-adaptive} is a model without classification layers, and the hidden state is initialized using Eq.(\ref{eq.phi_ada_noise}) during training and Eq.(\ref{eq.phi_ada}) during evaluation.
\end{itemize}

\subsection{Category Accuracy of Generated Texts}

\begin{table}[H]
	\centering
	\caption{Comparison of category accuracy of generated texts}
	\label{tab.accuaracy}
	\resizebox{0.6\textwidth}{!}{
		\begin{tabular}{ccccc}
			\toprule
			Method & MR & BR & AR & AVERAGE\\
			\midrule
			CatVRNN-nophi & 0.500 & 0.501 & 0.517 & 0.506\\
			VRNN-static & 0.528 & 0.594 & 0.968 & 0.697\\
			CatVRNN-static & 0.782 & \textbf{0.667} & \textbf{0.985} & 0.811\\
			VRNN-adaptive & 0.624 & 0.646 & 0.979 & 0.750\\
			CatVRNN-adaptive & \textbf{0.845} & 0.663 & 0.984 & \textbf{0.831}\\
			\midrule
			SentiGAN & 0.514 & 0.527 & 0.517 & 0.519\\
			CatGAN & 0.843 & 0.652 & \textbf{0.985} & 0.827\\
			\bottomrule
		\end{tabular}
	}
\end{table}

The results are displayed in Table \ref{tab.accuaracy}. To investigate whether the initialization of the hidden state can activate the generation state, comparisons were made with CatVRNN-nophi. The model randomly generates category texts without hidden state initialization during training, even if there is feedback from MTL. To investigate whether it is better to train multiple tasks than a single task, comparisons were carried out between the VRNN-static / adaptive and CatVRNN-static / adaptive models. It is interesting that a single generation task can generate a high-category accuracy sentence with only a special initial hidden state, and the MTL can improve the category accuracy, indicating that CatVRNN can directly learn the difference between category texts by activating different initial hidden states.

The comparison results in Table \ref{tab.accuaracy} indicate that the proposed model outperforms all other methods. The accuracy achieved by our model is highly promising, indicating that CatVRNN with MTL can better generate category text. 

\subsection{Quality of Generated Texts}

\begin{table}[H]
	\centering
	\caption{The perplexity of the models. Lower perplexity means better.}
	\label{tab.ppl}
	\resizebox{0.6\textwidth}{!}{
		\begin{tabular}{ccccc}
			\toprule
			Methods & MR & AR & BR & AVERAGE\\
			\midrule
			VRNN-static & 3.177 & 6.318 & 8.373 & 5.956\\
			CatVRNN-static & 2.724 & \textbf{3.619} & \textbf{5.280} & \textbf{3.874}\\
			VRNN-adaptive & 7.122 & 5.899 & 7.372 & 6.798\\
			CatVRNN-adaptive & \textbf{2.412} & 4.135 & 5.971 & 4.173\\
			\midrule
			SentiGAN & 4.028 & 12.122 & 40.869 & 19.006\\
			CatGAN & 9.893 & 38.152 & 26.722 & 24.922\\
			\bottomrule
		\end{tabular}
	}
\end{table}

Perplexity\cite{Brown1992} was further adopted to measure how well a model predicts a sample. The results are presented in Table \ref{tab.ppl}. Because CatVRNNs are trained entirely based on maximum likelihood estimation (MLE), their performance is naturally better than GANs trained based on unlikelihood, in terms of perplexity. After adversarial training, the current word distribution is more consistent with the discriminator, so the word distribution no longer matches the word distribution in the training set. These results are consistent with the results of the present study\cite{Wang2018}.

The harmonic average value of bilingual evaluation understudy (BLEU)\cite{Shi2018} was adopted to measure the quality of the generated texts in terms of fluency and diversity.

\begin{itemize}
	\item[$\bullet$] 
	\textbf{Forward BLEU} ($BLEU_{F}$) uses the training set as a reference and evaluates each generated text using the BLEU score. It measures the precision (fluency) of the generator.	
	\item[$\bullet$] 
	\textbf{Backward BLEU} ($BLEU_{B}$) uses the generated text as a reference and evaluates each text in the training set using the BLEU score. It measures the recall (diversity) of the generator.	
	\item[$\bullet$]
	\textbf{Harmonic BLEU} ($BLEU_{HA}$) is the harmonic average value of $BLEU_{F}$ and $BLEU_{B}$, and is defined by Eq.(\ref{eq.bleu_ha}).	  	
\end{itemize}

\begin{equation}
	BLEU_{HA}=\frac{2 \times BLEU_{F} \times BLEU_{B}}{BLEU_{F}+BLEU_{B}}
	\label{eq.bleu_ha}
\end{equation}

The results are shown in Table \ref{tab.BLUE_MR}, \ref{tab.BLUE_BR}, \ref{tab.BLUE_AR} and \ref{tab.quality}. Comparing VRNN-static / adaptive and CatVRNN-static / adaptive, MTL improves not only the category accuracy of the generated text, but also the quality of the generated text. The CatVRNNs improve the fluency and diversity of the generated text by back propagation of the classification task. This indicates that the classification task can help the generation task to learn a more detailed distribution of tokens.

\begin{table}[H]
	\centering
	\caption{Comparison of quality of generated texts on MR}
	\label{tab.BLUE_MR}
	\resizebox{0.6\textwidth}{!}{
		\begin{tabular}{c|cccc|cc}
			\toprule
			Metrics & \tabincell{c}{VRNN-\\static} & \tabincell{c}{VRNN-\\adaptive} & \tabincell{c}{CatVRNN-\\static} & \tabincell{c}{CatVRNN-\\adaptive} & SentiGAN & CatGAN\\
			\midrule
			$BLEU_{F}$-2 & 0.802 & 0.789 & 0.828 & 0.807 & \textbf{0.896} & 0.882\\
			$BLEU_{F}$-3 & 0.618 & 0.611 & 0.670 & 0.647 & \textbf{0.747} & 0.670\\
			$BLEU_{F}$-4 & 0.496 & 0.493 & 0.559 & 0.539 & \textbf{0.589} & 0.445\\
			$BLEU_{F}$-5 & 0.422 & 0.424 & \textbf{0.489} & 0.473 & 0.466 & 0.292\\
			\midrule 			
			$BLEU_{B}$-2 & 0.830 & 0.819 & \textbf{0.833} & 0.807 & 0.721 & 0.825\\
			$BLEU_{B}$-3 & 0.665 & 0.647 & \textbf{0.681} & 0.646 & 0.588 & 0.680\\
			$BLEU_{B}$-4 & 0.545 & 0.524 & \textbf{0.568} & 0.531 & 0.467 & 0.563\\
			$BLEU_{B}$-5 & 0.468 & 0.448 & \textbf{0.495} & 0.459 & 0.380 & 0.477\\
			\midrule
			$BLEU_{HA}$-2 & 0.816 & 0.804 & 0.830 & 0.807 & 0.799 & \textbf{0.853}\\
			$BLEU_{HA}$-3 & 0.641 & 0.628 & \textbf{0.675} & 0.646 & 0.658 & \textbf{0.675}\\
			$BLEU_{HA}$-4 & 0.519 & 0.508 & \textbf{0.563} & 0.535 & 0.521 & 0.497\\
			$BLEU_{HA}$-5 & 0.444 & 0.436 & \textbf{0.492} & 0.466 & 0.419 & 0.362\\
			\bottomrule
		\end{tabular}
	}
\end{table}

\begin{table}[H]
	\centering
	\caption{Comparison of quality of generated texts on BR}
	\label{tab.BLUE_BR}
	\resizebox{0.6\textwidth}{!}{
		\begin{tabular}{c|cccc|cc}
			\toprule
			Metrics & \tabincell{c}{VRNN-\\static} & \tabincell{c}{VRNN-\\adaptive} & \tabincell{c}{CatVRNN-\\static} & \tabincell{c}{CatVRNN-\\adaptive} & SentiGAN & CatGAN\\
			\midrule
			$BLEU_{F}$-2 & 0.894 & 0.859 & 0.894 & 0.879 & 0.873 & \textbf{0.932}\\
			$BLEU_{F}$-3 & 0.745 & 0.698 & 0.761 & 0.738 & 0.678 & \textbf{0.810}\\
			$BLEU_{F}$-4 & 0.572 & 0.537 & 0.610 & 0.590 & 0.446 & \textbf{0.638}\\
			$BLEU_{F}$-5 & 0.428 & 0.416 & 0.484 & 0.473 & 0.264 & \textbf{0.513}\\
			\midrule
			$BLEU_{B}$-2 & \textbf{0.877} & 0.867 & \textbf{0.877} & 0.876 & 0.783 & 0.833\\
			$BLEU_{B}$-3 & \textbf{0.731} & 0.708 & \textbf{0.731} & 0.723 & 0.586 & 0.690\\
			$BLEU_{B}$-4 & 0.580 & 0.553 & \textbf{0.583} & 0.572 & 0.375 & 0.552\\
			$BLEU_{B}$-5 & 0.462 & 0.438 & \textbf{0.468} & 0.456 & 0.220 & 0.444\\
			\midrule
			$BLEU_{HA}$-2 & \textbf{0.885} & 0.863 & \textbf{0.885} & 0.877 & 0.826 & 0.880\\
			$BLEU_{HA}$-3 & 0.738 &	0.703 &	\textbf{0.746} & 0.730 & 0.629 & 0.745\\
			$BLEU_{HA}$-4 & 0.576 &	0.545 &	\textbf{0.596} & 0.581 & 0.407 & 0.592\\
			$BLEU_{HA}$-5 & 0.444 &	0.427 &	\textbf{0.476} & 0.464 & 0.240 & 0.476\\
			\bottomrule
		\end{tabular}
	}
\end{table}

\begin{table}[H]
	\centering
	\caption{Comparison of quality of generated texts on AR}
	\label{tab.BLUE_AR}
	\resizebox{0.6\textwidth}{!}{
		\begin{tabular}{c|cccc|cc}
			\toprule
			Metrics & \tabincell{c}{VRNN-\\static} & \tabincell{c}{VRNN-\\adaptive} & \tabincell{c}{CatVRNN-\\static} & \tabincell{c}{CatVRNN-\\adaptive} & SentiGAN & CatGAN\\
			\midrule
			$BLEU_{F}$-2 & 0.915 & 0.921 & 0.920 & 0.927 & 0.896 & \textbf{0.958}\\
			$BLEU_{F}$-3 & 0.794 & 0.820 & 0.810 & 0.843 & 0.738 & \textbf{0.891}\\
			$BLEU_{F}$-4 & 0.644 & 0.694 & 0.671 & 0.738 & 0.542 & \textbf{0.782}\\
			$BLEU_{F}$-5 & 0.509 & 0.584 & 0.536 & 0.638 & 0.359 & \textbf{0.678}\\
			\midrule
			$BLEU_{B}$-2 & 0.919 & 0.903 & \textbf{0.921} & 0.910 & 0.813 & 0.858\\
			$BLEU_{B}$-3 & 0.790 & 0.768 & \textbf{0.791} & 0.776 & 0.620 & 0.725\\
			$BLEU_{B}$-4 & 0.639 & 0.619 & \textbf{0.645} & 0.626 & 0.411 & 0.588\\
			$BLEU_{B}$-5 & 0.505 & 0.492 & \textbf{0.514} & 0.497 & 0.253 & 0.469\\
			\midrule
			$BLEU_{HA}$-2 & 0.917 & 0.912 & \textbf{0.920} & 0.918 & 0.852 & 0.905\\
			$BLEU_{HA}$-3 & 0.792 & 0.793 & 0.800 & \textbf{0.808} & 0.674 & 0.799\\
			$BLEU_{HA}$-4 & 0.641 & 0.654 & 0.658 & \textbf{0.677} & 0.467 & 0.671\\
			$BLEU_{HA}$-5 & 0.507 & 0.534 & 0.525 & \textbf{0.559} & 0.297 & 0.554\\
			\bottomrule
		\end{tabular}
	}	
\end{table}

\begin{table}[H]
	\centering
	\caption{The average quality of generated texts across all datasets}
	\label{tab.quality}
	\resizebox{0.6\textwidth}{!}{
		\begin{tabular}{c|cccc|cc}
			\toprule
			Metrics & \tabincell{c}{VRNN-\\static} & \tabincell{c}{VRNN-\\adaptive} & \tabincell{c}{CatVRNN-\\static} & \tabincell{c}{CatVRNN-\\adaptive} & SentiGAN & CatGAN\\
			\midrule
			$BLEU_{F}$-2 & 0.870 & 0.856 & 0.881 & 0.871 & 0.888 & \textbf{0.924} \\
			$BLEU_{F}$-3 & 0.719 & 0.710 & 0.747 & 0.743 & 0.721 & \textbf{0.790} \\
			$BLEU_{F}$-4 & 0.571 & 0.575 & 0.613 & \textbf{0.622} & 0.526 & \textbf{0.622} \\
			$BLEU_{F}$-5 & 0.453 & 0.475 & 0.503 & \textbf{0.528} & 0.363 & 0.494 \\
			\midrule 			
			$BLEU_{B}$-2 & 0.875 & 0.863 & \textbf{0.877} & 0.864 & 0.772 & 0.839 \\
			$BLEU_{B}$-3 & 0.729 & 0.708 & \textbf{0.734} & 0.715 & 0.598 & 0.698 \\
			$BLEU_{B}$-4 & 0.588 & 0.565 & \textbf{0.599} & 0.576 & 0.418 & 0.568 \\
			$BLEU_{B}$-5 & 0.478 & 0.459 & \textbf{0.492} & 0.471 & 0.284 & 0.463 \\
			\midrule
			$BLEU_{HA}$-2 & 0.873 & 0.860 & \textbf{0.879} & 0.868 & 0.826 & \textbf{0.879} \\
			$BLEU_{HA}$-3 & 0.724 & 0.709 & \textbf{0.741} & 0.729 & 0.654 & \textbf{0.741} \\
			$BLEU_{HA}$-4 & 0.579 & 0.570 & \textbf{0.606} & 0.598 & 0.465 & 0.593 \\
			$BLEU_{HA}$-5 & 0.465 & 0.467 & \textbf{0.498} & \textbf{0.498} & 0.319 & 0.478 \\
			\bottomrule
		\end{tabular}
	}
\end{table}

From the comparison between CatVRNNs and baselines, CatVRNNs can generate more diverse texts by losing a small amount of text fluency. The difference in $BLEU_{B}$ is larger than the difference in $BLEU_{F}$ between CatVRNNs and GANs; therefore, the quality of texts generated by CatVRNNs is slightly better. However, GANs adopt different methods to improve diversity. In comparing, the diversity of texts generated by GANs and CatVRNNs, GANs are still lacking. In text generation tasks, the problem of mode collapse of GANs can only be improved rather than solved, proving that texts generated by GANs have good fluency but lack diversity. 

\subsection{Comparison between CatVRNN-adaptive and CatVRNN-static}

\begin{figure}[H]
	\centering
	\subfigure[global loss curves]{
		\includegraphics[width=0.45\textwidth]{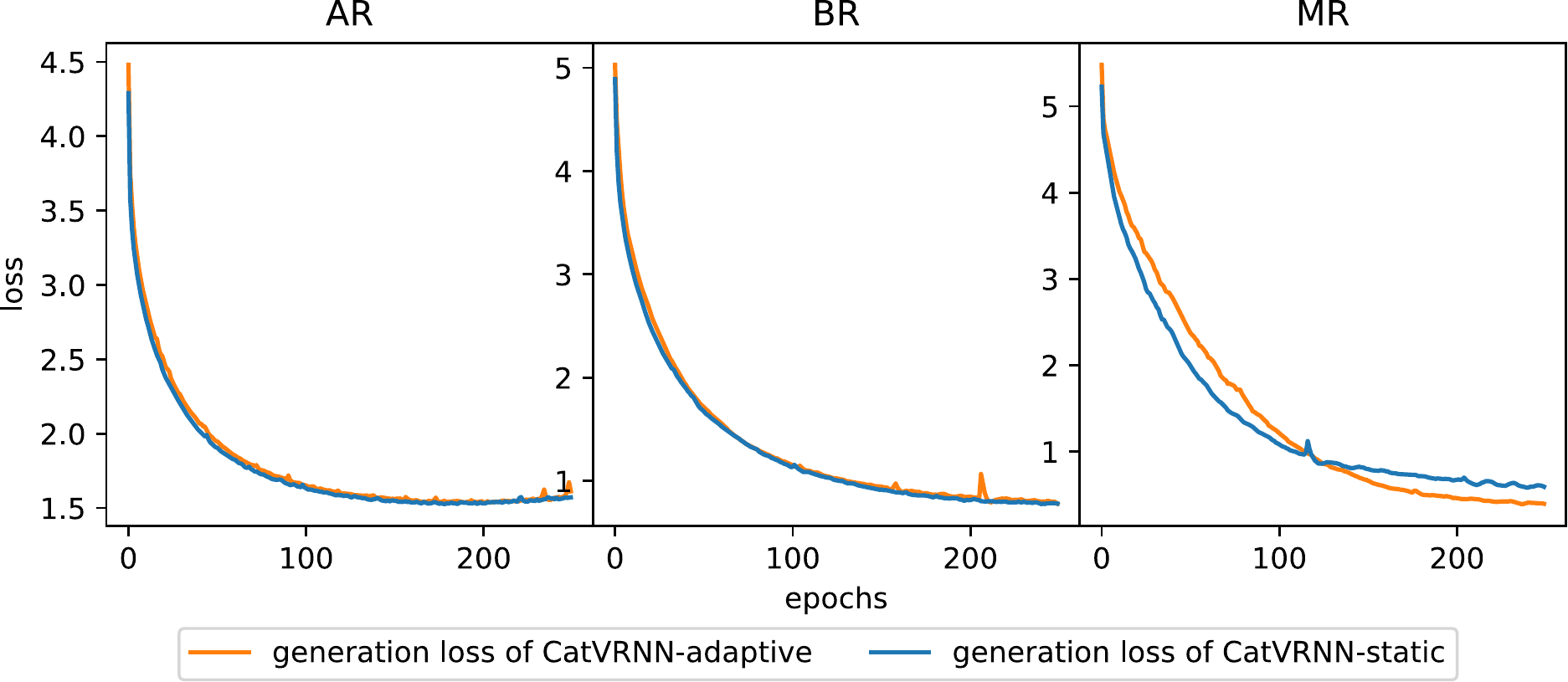}
	}
	\subfigure[local loss curves]{
		\includegraphics[width=0.45\textwidth]{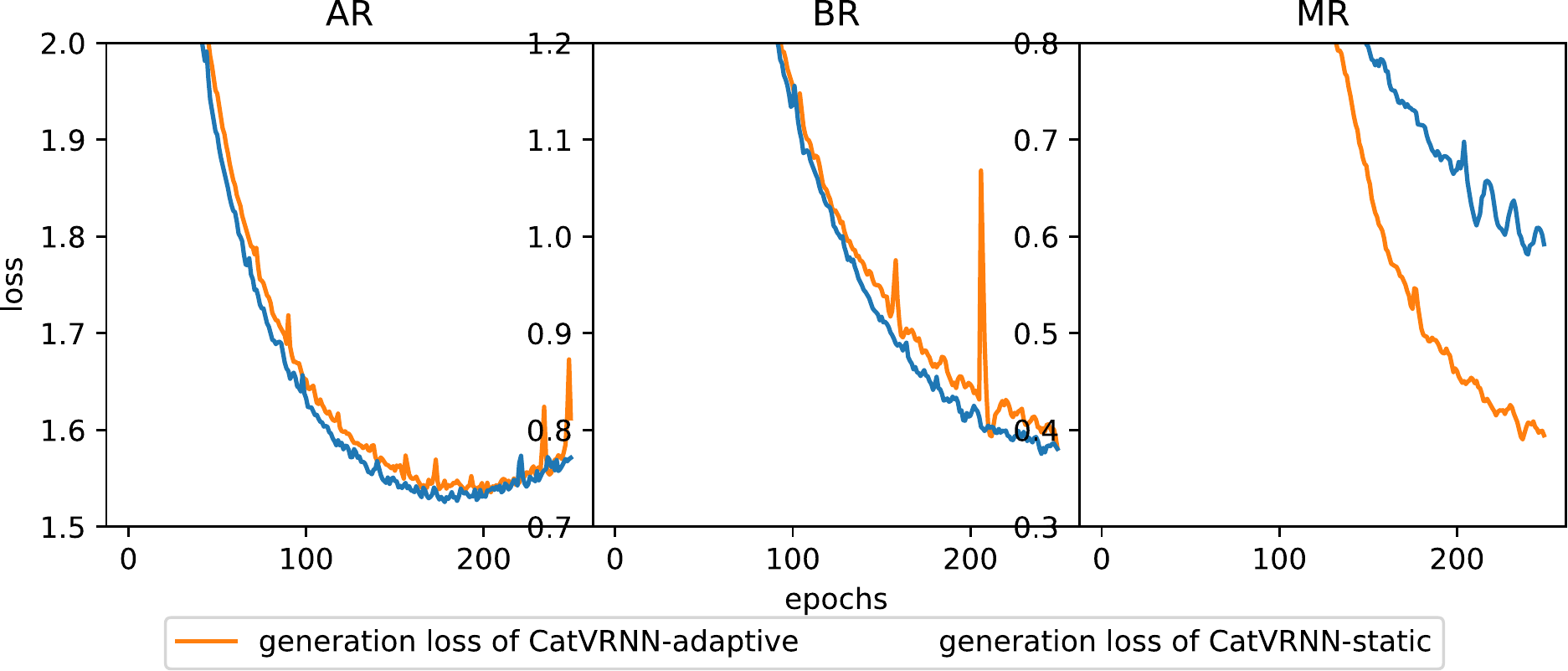}
	}
	\caption{The loss curves of CatVRNN with different functions of hidden state initialization. The curves in the (b) figure are an intercept and enlargement of the tail of the curves in the (a) figure.}
	\label{fig.h_initialization}
\end{figure}

As shown in Fig.\ref{fig.h_initialization}, on a global scale, the loss curves of CatVRNN-adaptive and CatVRNN-static differ by little, although CatVRNN-adaptive causes greater fluctuations at the end of training. This is the direct result of the negative effects of adaptive initialization. The underlying cause of this problem is that the initialization value of the hidden state $h_{0}$ has a large impact on CatVRNN. With the static initialization function $\phi_{static}$, $h_{0}$ remains constant. The effect of $h_{0}$ on the loss is gradually minimized as the training proceeds, allowing CatVRNN to focus on the generation task. However, with the adaptive initialization function $\phi_{adaptive}$, $h_{0}$ changes as the gradient progresses. The gradient update does not always move towards the optimal solution and can produce loss fluctuations. The fluctuation of loss is magnified if the model is sensitive to the $h_{0}$. 

As shown in Table \ref{tab.accuaracy}, although the category accuracy of texts generated by CatVRNN-static is better than that of CatVRNN-adaptive, the overall category accuracy of texts generated by CatVRNN-adaptive is better than those of CatVRNN-static and the baselines.

As shown in Table \ref{tab.BLUE_MR}, \ref{tab.BLUE_BR} and \ref{tab.BLUE_AR}, in comparing the generated text quality, CatVRNN-static had advantages. To generate good quality text, numerous attempts were made to determine the hyperparameter $\omega$; $h_{0}$ calculated with different $\omega$s, affects the category accuracy and quality of the generated texts. It is difficult to find a decent $\omega$ that can balance all indicators. In addition,  there is no better static approach to determine $\omega$ other than trial-and-error exploration. This was one motivation for us in proposing CatVRNN-adaptive, which adaptively determines $\omega$ by the model itself. As shown in Table \ref{tab.quality}, the difference between CatVRNN-adaptive and CatVRNN-static is acceptable when comparing using $BLEU_{HA}$ with respect to the overall quality. The largest difference between the two methods was merely 0.012($BLEU_{HA}$-3). This is also the biggest difference between the CatVRNN-adaptive and the baselines. In $BLEU_{HA}$-4 and $BLEU_{HA}$-5, CatVRNN-adaptive is also superior to the baselines.

At the expense of sacrificing a little quality in the generated text, CatVRNN-adaptive provides greater advantages because it does not require determining hyperparameters, has superior scalability in generating more categories of text and can generate texts with higher category accuracy.

\subsection{Impact of KL term and feature extractors}

\begin{figure}[H]
	\centering
	\subfigure[CatVRNN-static]{
		\includegraphics[width=0.25\textwidth]{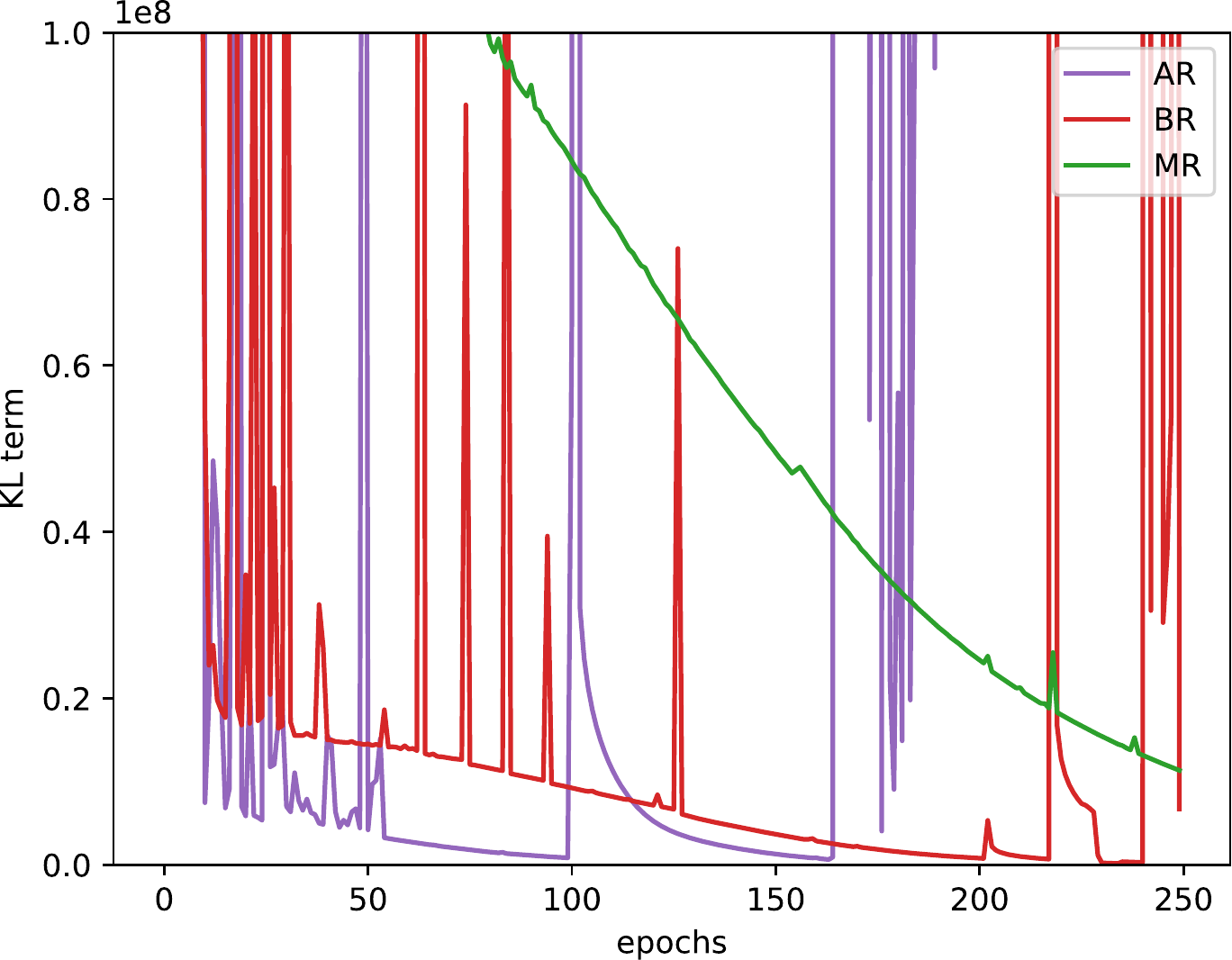} 
	}
	\subfigure[CatVRNN-adaptive]{
		\includegraphics[width=0.25\textwidth]{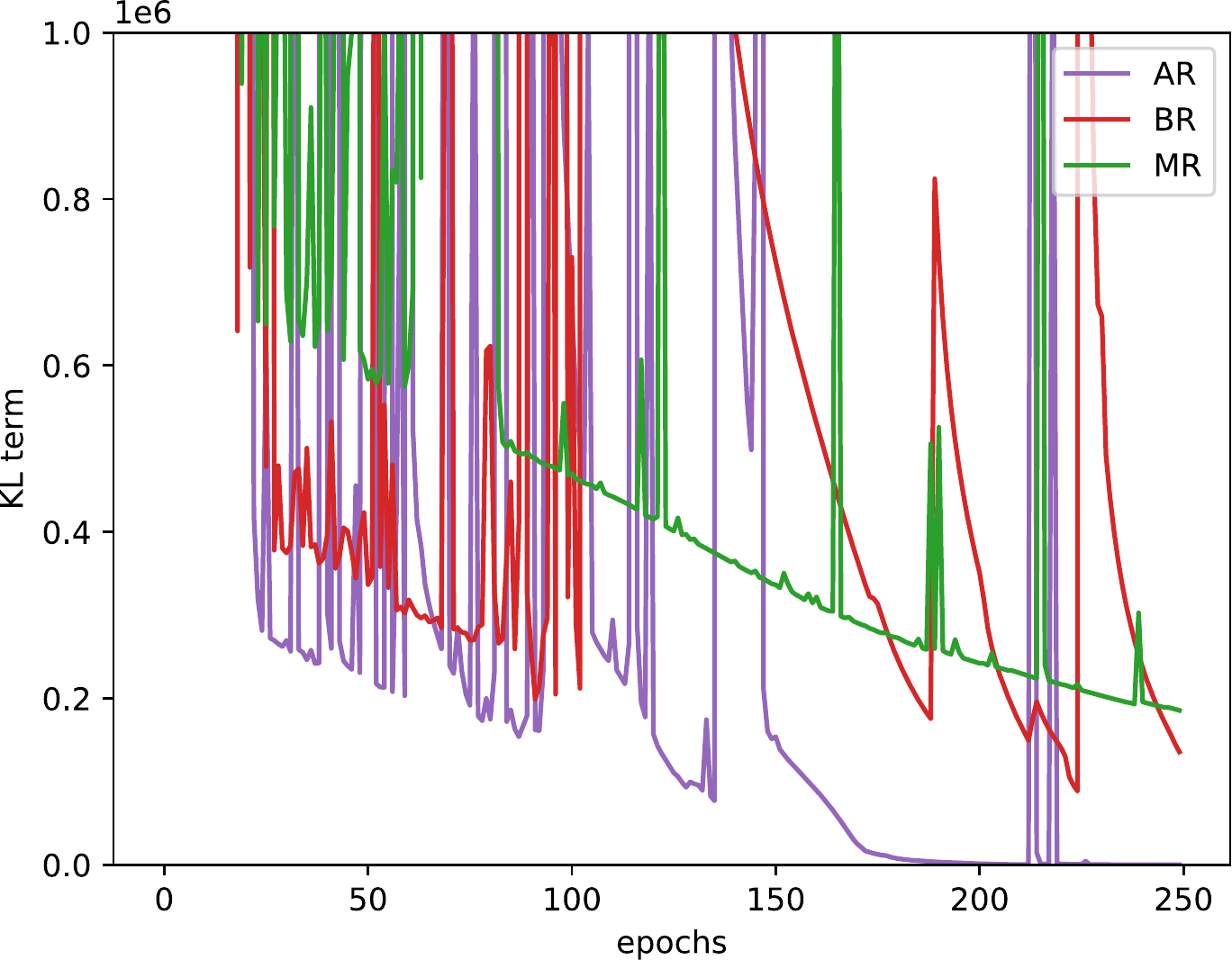} 
	}
	\caption{The curves of KL term. Different colored lines represent the loss on various datasets. It is very crucial to remember that the ordinate units of the pictures are $1e8$ and $1e6$, respectively. Since the curves fluctuate so much, the most concentrated part of the results were selected to show.}
	\label{fig.kld_curves}
\end{figure}

\begin{figure}[H]
	\centering
	\subfigure[CatVRNN-static]{
		\includegraphics[width=0.45\linewidth]{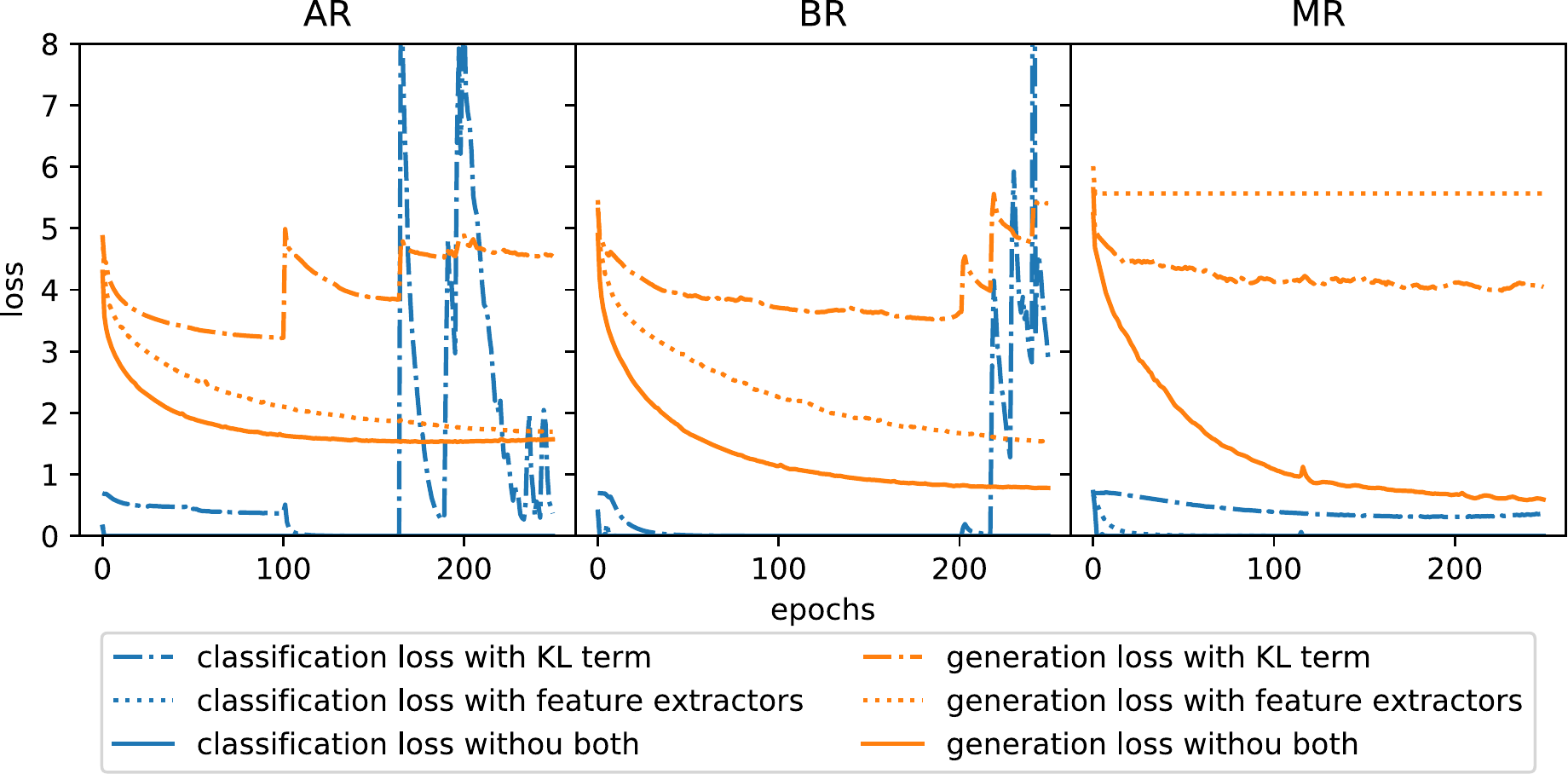} 
	}
	\subfigure[CatVRNN-adaptive]{
		\includegraphics[width=0.45\linewidth]{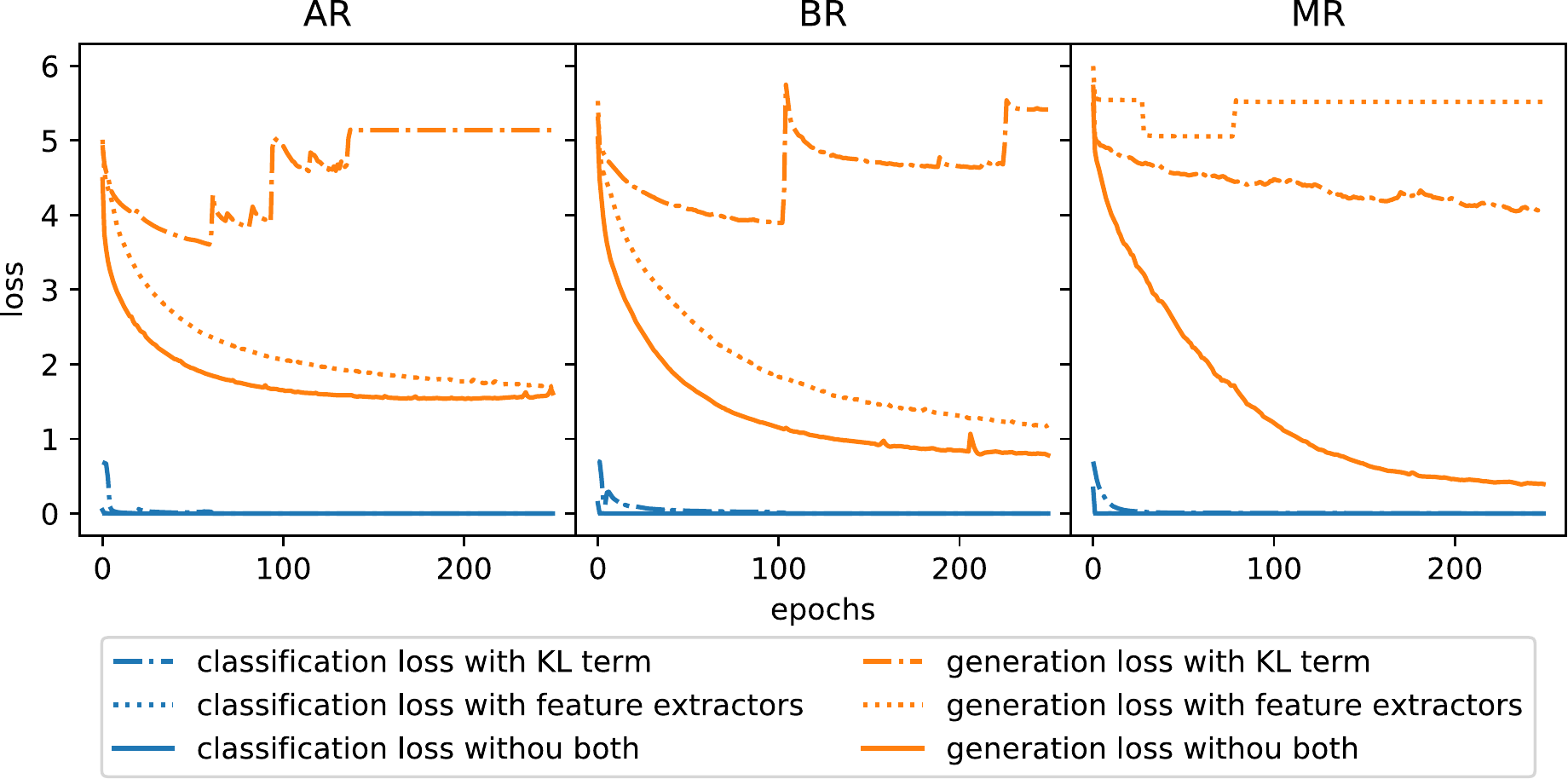} 
	}
	\caption{The loss curves of the generation task and classification task on various datasets using KL term or feature extractors. Loss with KL term means the models employ KL term during training; loss with feature extractors means the models employ feature extractors during training; loss without both means the models neither employ KL term nor feature extractors during training. Because the classification loss with feature extractors coincides with the classification loss without both, the graphic display is obscured.}
	\label{fig.kld_impact_loss}
\end{figure}

After the KL term is introduced in the training of CatVRNN, the curves of the KL term are shown in Fig.\ref{fig.kld_curves}. The curves have undergone significant swings, and they are particularly slow to converge. Because of the powerful noise caused by the huge KL term, not only do the generation tasks converge badly, but some classification tasks fail to converge, as shown in Fig.\ref{fig.kld_impact_loss}. The impact of the KL term on the loss of the generation task is particularly noticeable. Excessive loss indicates that the generated text is unreadable.

In addition, as displayed in Table \ref{tab.BLUE_MR}, \ref{tab.BLUE_BR}, and \ref{tab.BLUE_AR}, $BLEU_{F}$ and $BLEU_{B}$ both indicate that the models without the KL term still introduces the noise though $z$. $BLEU_{F}$ illustrates that CatVRNNs do not clone the samples to generate the same texts as the samples; $BLEU_{B}$ illustrates that CatVRNNs do not generate the same texts repeatedly.

The loss of generation tasks is also significantly influenced by feature extractors. In CatVRNN, feature extractors are initially used to extract features from a word. A word is a semantic unit in text. As a result, there are no features to extract from a word. Forced feature extraction definitely results in word meaning confusion, which is bound to have a detrimental influence on the loss of generation tasks.

\subsection{Resource Consumption}

There is one generator in CatVRNN, and the classifier and generator share most of the parameters; therefore, the CatVRNN's parameters are fewer. CatVRNN simply employs MLE training rather than adversarial training, resulting in a lower training complexity and shorter training time. Compared to CatVRNN-static, CatVRNN-adaptive has an additional linear layer because of the initialization function $\phi_{adaptive}$ of the hidden state, so CatVRNN-adaptive has a few more parameters and is associated with longer processing times. Detailed information is provided in Table \ref{tab.consumption}.

\begin{table}[H]
	\centering
	\caption{Resource consumption. $|\nu|$ is the size of vocabulary. For GANs, it just counts the generator's parameters, and the time consumed during adversarial training.}
	\label{tab.consumption}
	\resizebox{0.6\textwidth}{!}{
		\begin{tabular}{ccc}
			\toprule
			Methods & parameters & time consumption per iteration($s$)\\
			\midrule
			CatVRNN-static & \textbf{1251674+601$|\nu|$} & \textbf{0.332} \\
			CatVRNN-adaptive & 1559130+601$|\nu|$ & 0.340 \\
			SentiGAN & 6273363+1064$|\nu|$ & 231.962 \\
			CatGAN & 8453122+1057$|\nu|$ & 48.156\\
			\bottomrule
		\end{tabular}
	}
\end{table}	

\section{Discussion}

\begin{figure}[htbp]
	\centering
	\subfigure[on Mr]{
		\includegraphics[width=1.5in]{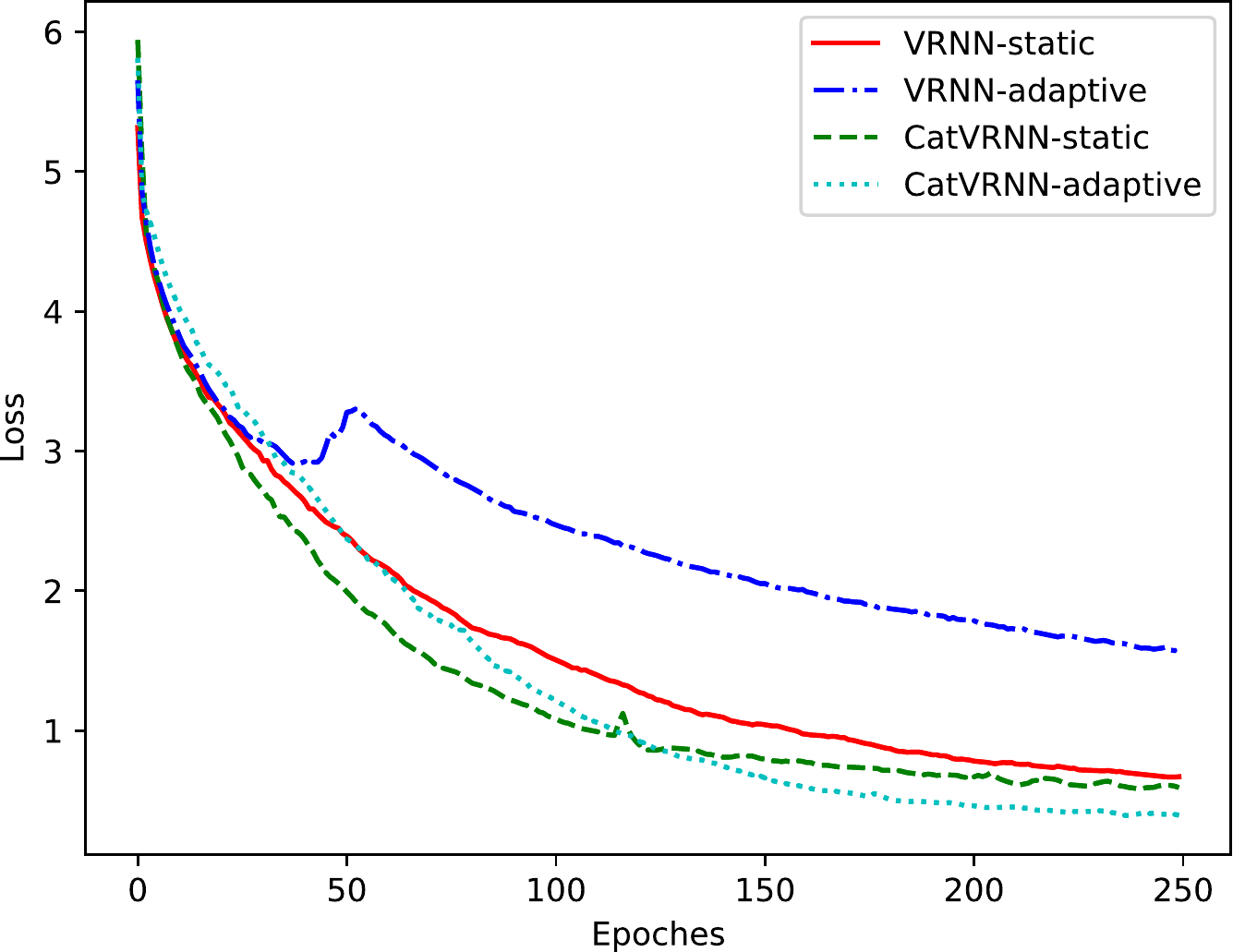}
	}
	\subfigure[on Br]{
		\includegraphics[width=1.5in]{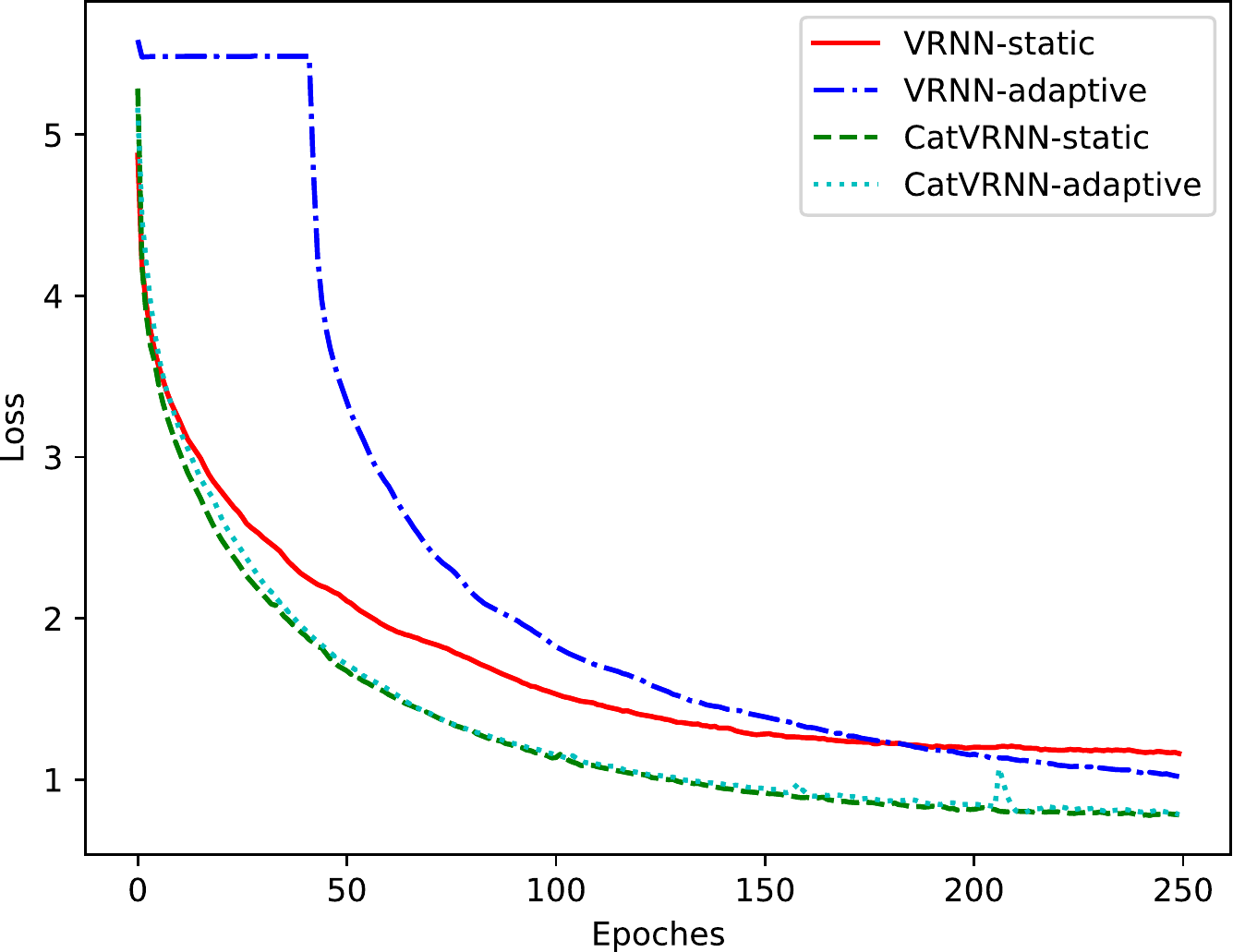}
	}
	\subfigure[on Ar]{
		\includegraphics[width=1.5in]{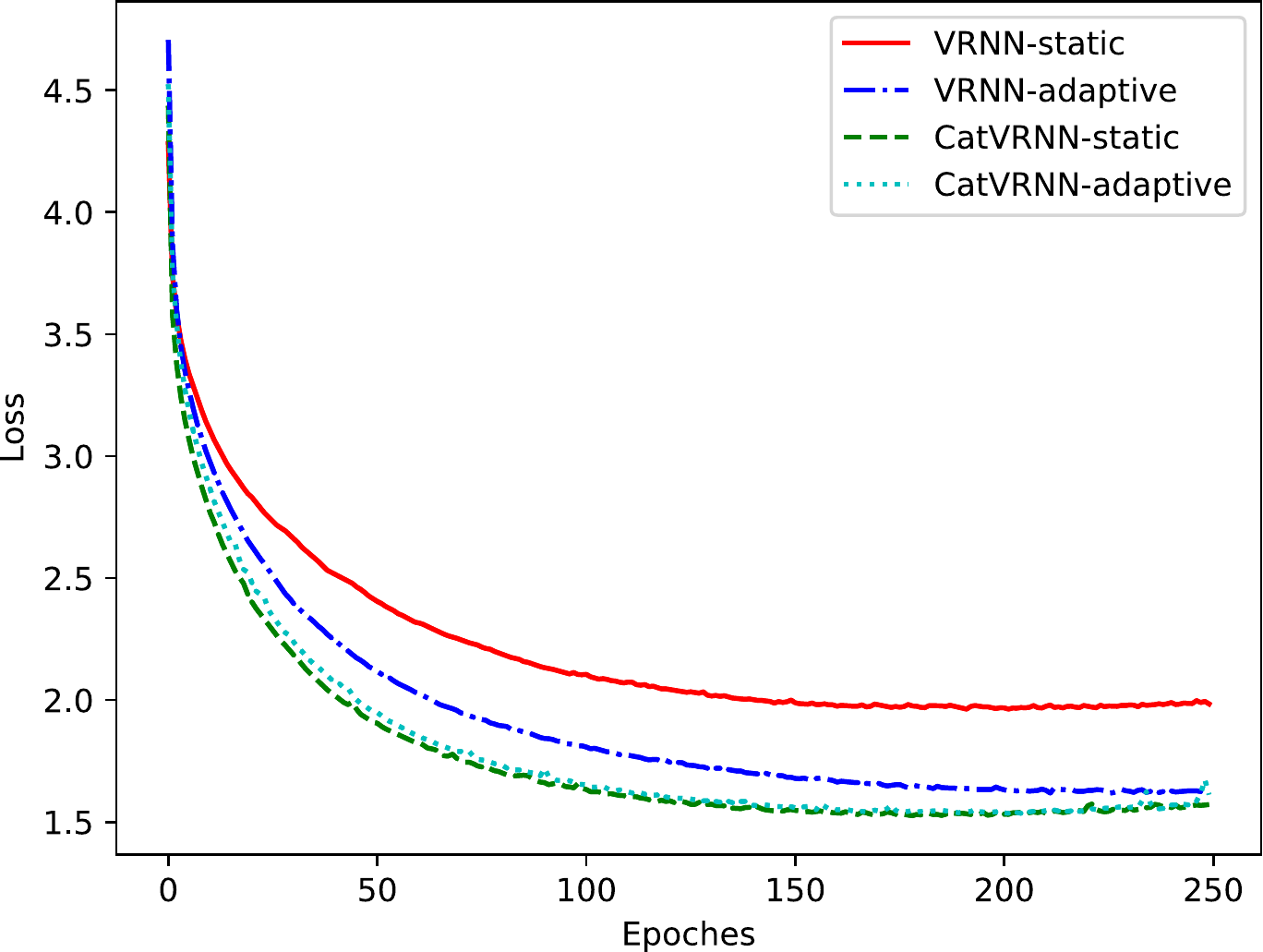}
	}
	\caption{The illustration of learning curves. When $C$ is set to 2, the training losses of CatVRNNs are lower than that of VRNNs. This indicates that the classification task has a positive impact on the generation task, resulting in further loss reduction and better fluency of generated texts.}
	\label{fig.curves}
\end{figure}

The learning curves of the CatVRNNs are shown in Fig.\ref{fig.curves}. The category accuracy and quality of the generated texts indicate that CatVRNNs benefit from MTL. Without the classification task, convergence is not smooth in CatVRNNs. Can the performance of the generation task be improved by increasing the difficulty of the classification task? This problem will be illustrated in terms of the quality and category accuracy of the generated texts. 

\subsection{Quality of Generated Texts Impacted by Classification}

To illustrate the impact of classification on the quality of generated texts, a new dataset named the impact of classification on quality (ICQ-base) was built comprising five types of products (cell phone and accessories, grocery and gourmet food, baby, beauty, and pet supplies) from Amazon reviews\cite{McAuley2015}. Each product has two sentiment classes (negative and positive). This dataset can be transformed into four datasets using different combinations. In general, the difficulty of the classification task can be increased by increasing the categories. The vocabulary and the samples of the four deformed datasets are all the same and retain the quality of texts generated by CatVRNN, only being affected by the difficulty of classification. 

\begin{figure}[h]
	\centering
	\includegraphics[width=2.5in]{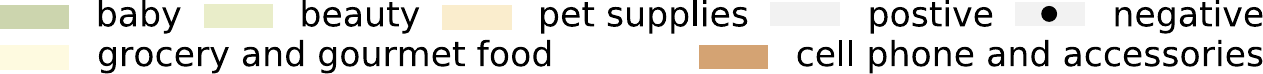}
	
	\subfigure[ICQ-1C]{
		\includegraphics[width=1in]{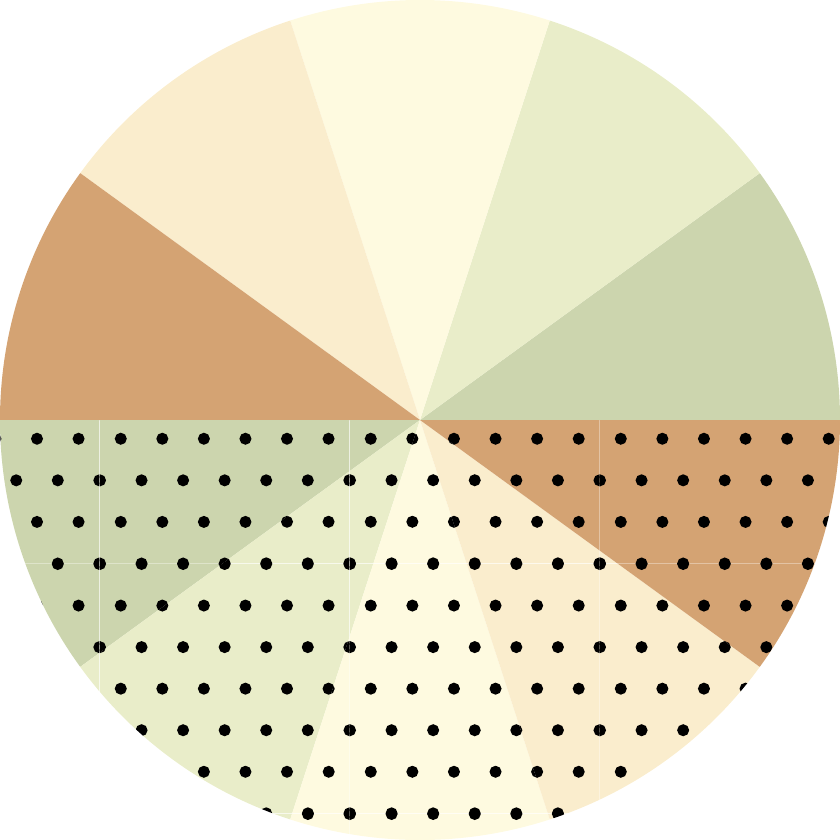}
		\label{fig.ICQ-1C}
	}
	\subfigure[ICQ-2C]{
		\includegraphics[width=1in]{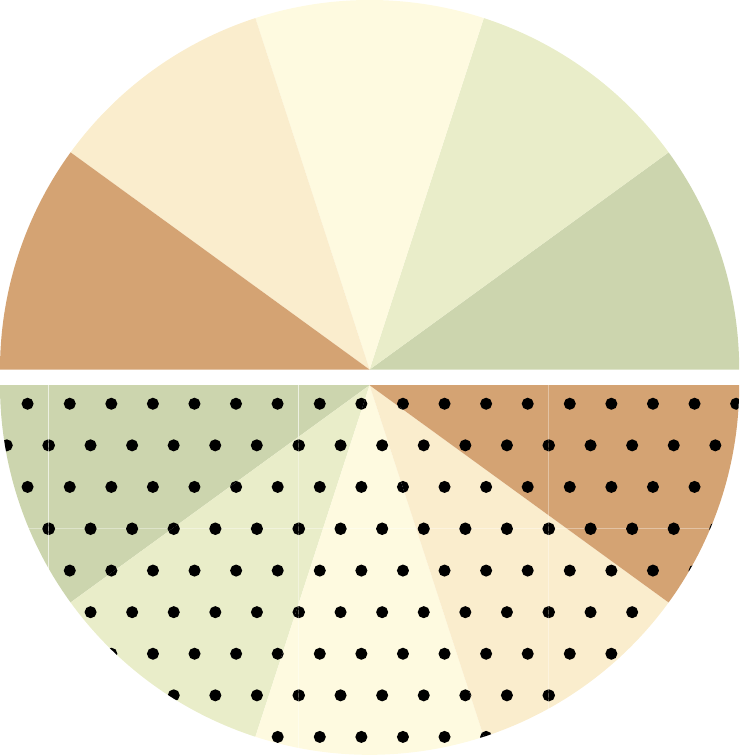}
		\label{fig.ICQ-2C}
	}
	\subfigure[ICQ-5C]{
		\includegraphics[width=1in]{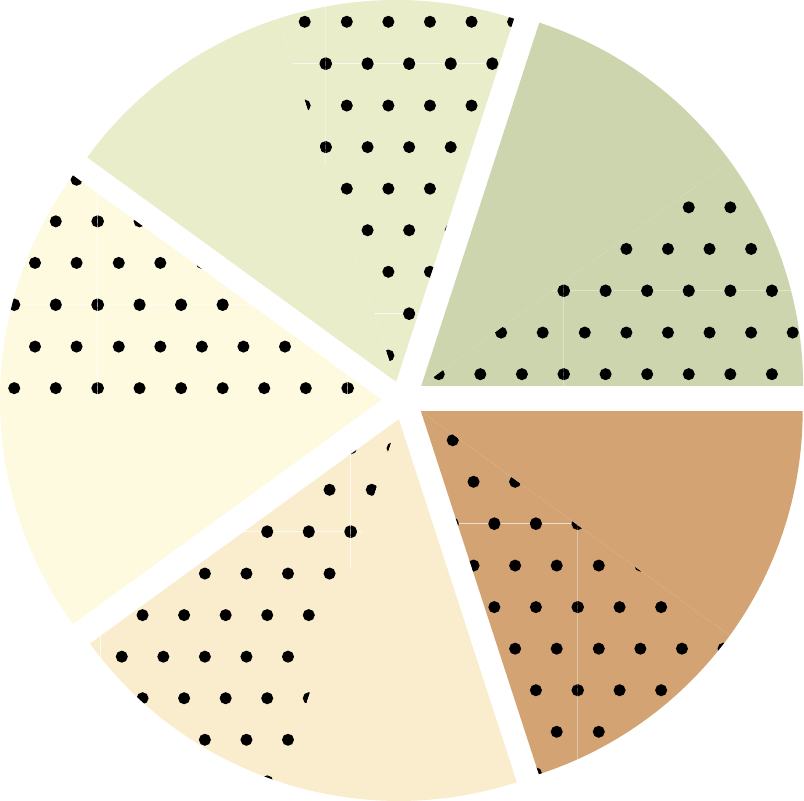}
		\label{fig.ICQ-5C}
	}
	\subfigure[ICQ-10C]{
		\includegraphics[width=1in]{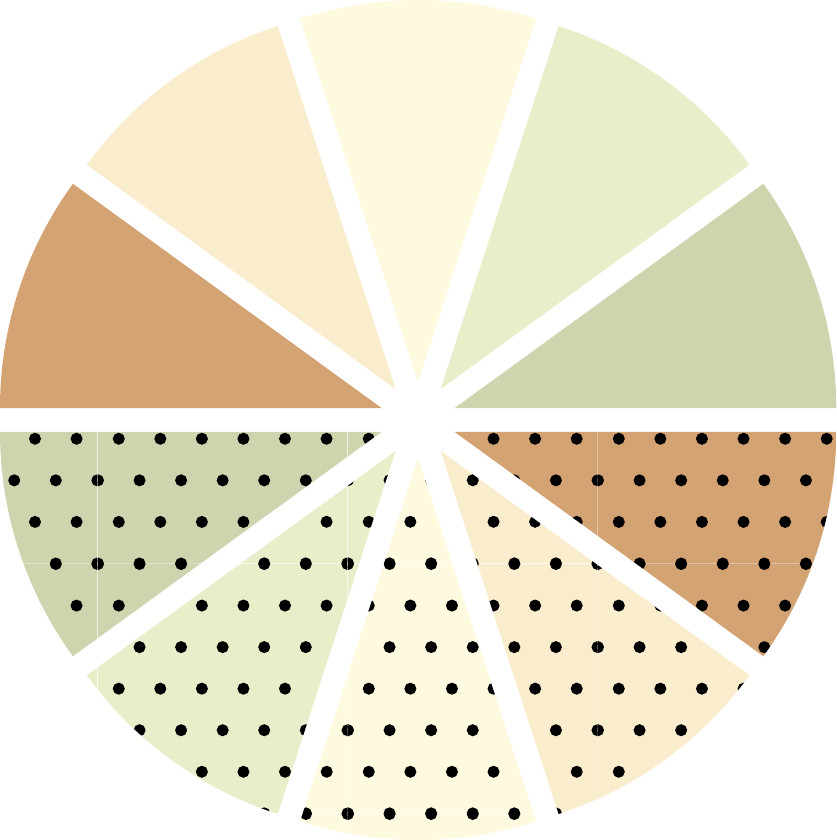}
		\label{fig.ICQ-10C}
	}
	\caption{Composition of ICQ-base deformation. Sectors of the same color are the same product; Pure colors are samples of positive; Patterned are samples of negative; White gaps between sectors are category dividing lines; And the size of samples represented by a sector is 1000.}
	\label{fig.ICQ}
\end{figure}

As shown in Fig.\ref{fig.ICQ}, the ICQ-base dataset is transformed into four datasets using different combinations. The detailed configurations are as follows:

\textbf{ICQ-1C} They are all product reviews, regardless of the type of product or sentiment. The ICQ-base may be considered as a category.    

\textbf{ICQ-2C} Only product review sentiment factors are considered. The ICQ-base can be separated into two categories.    

\textbf{ICQ-5C} Only product category is considered. The ICQ-base can be divided into five categories.

\textbf{ICQ-10C} Even after considering the product category and the product review sentiment factors, reviews of particular products with distinct sentiments are treated as a category. The ICQ-base may be divided into ten categories.

The CatVRNN-adaptive model was adopted to train datasets except ICQ-1C. CatVRNN-adaptive can fit the learning curve of CatVRNN-static well, and CatVRNN-adaptive supports tasks with multiple categories. For a fair comparison, VRNN-adaptive model was adopted to train on ICQ-1C. The difficulty of the classification task of VRNN-adaptive on ICQ-1C is defined as none.

\begin{figure}[H]
	\centering
	\subfigure[generation tasks]{
		\includegraphics[width=2in]{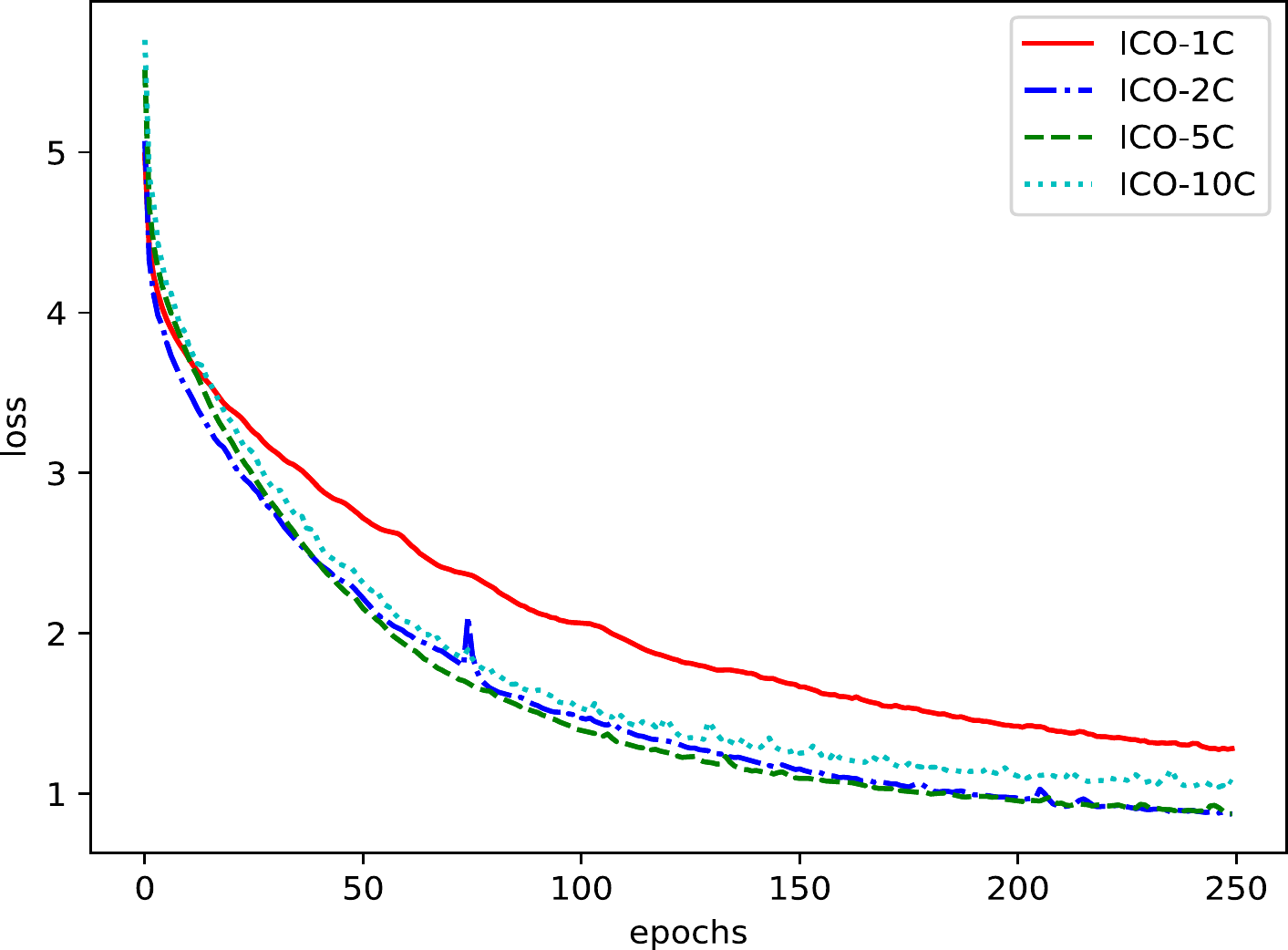}
		\label{fig.ar5_all}
	}
	\subfigure[classification tasks]{
		\includegraphics[width=2in]{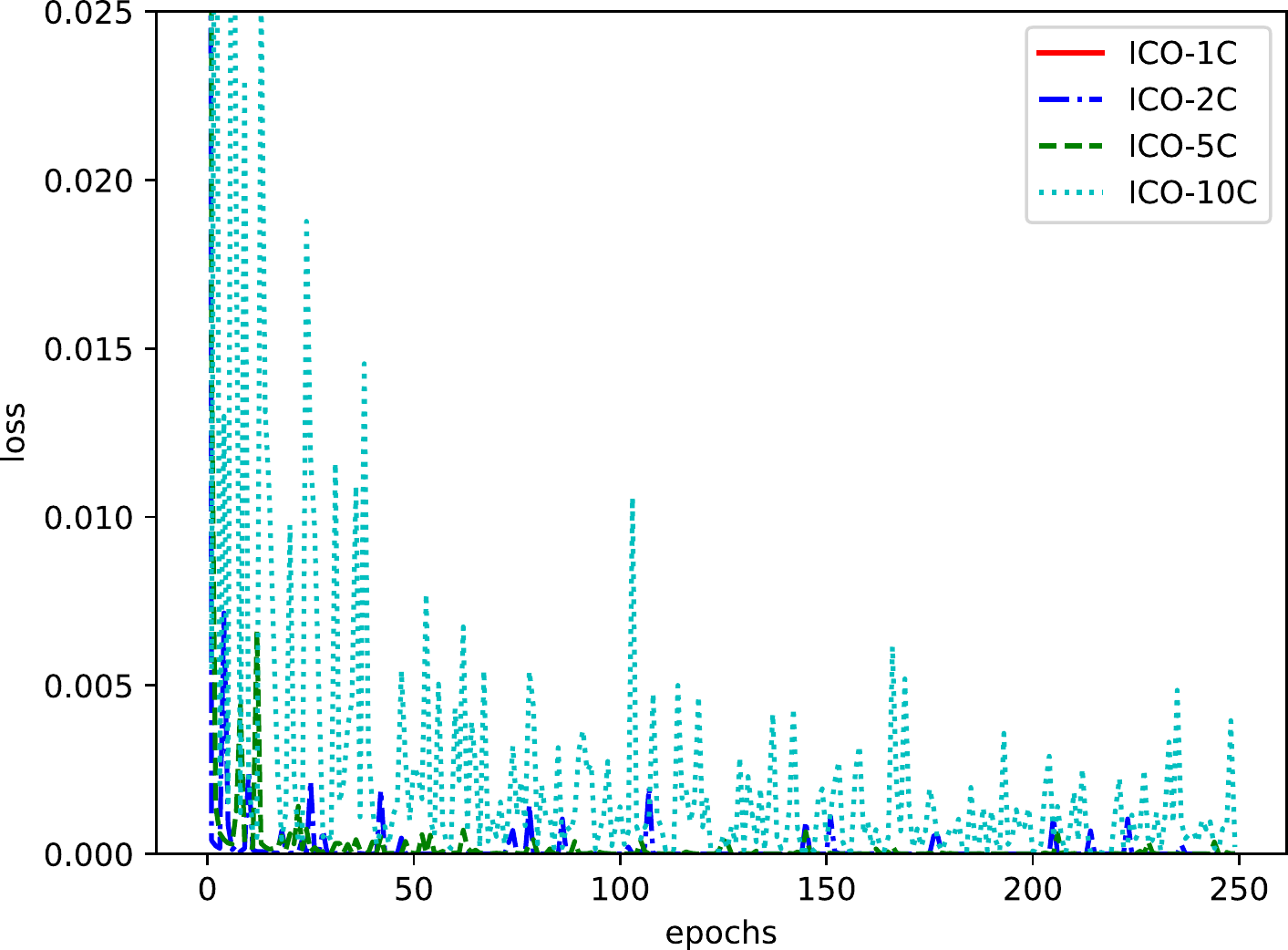}
		\label{fig.ar5_classification}
	}
	\caption{The impact of classification difficulty on learning curves. Compared to the generation task, the loss in classification task converges faster and fluctuates in a lower region. Even so, the loss of the classification task increases as the difficulty of the classification task increases.}
	\label{fig.ar5}
\end{figure}

As shown in Fig.\ref{fig.ar5}, the loss is improved significantly when the categories of the classification task are in the range of 0 to 2. When the categories of the classification task are from 2 to 5, the loss is almost unchanged. The loss is worse when the categories of the classification task is within the range of 5 to 10. As shown in Fig.\ref{fig.ar5_classification}, classification is difficult for the CatVRNN-adaptive model on ICQ-10C. The learning curve of the classification on ICQ-10C cannot converge to 0. However, the learning curves of classification on ICQ-2C and ICQ-5C converge to 0 during the early phase of training. The classification tasks that are beyond the ability of the model cause the model not to focus on the generation task. This means that the quality of the generation task can be improved by increasing the difficulty of the classification task, but this process has an upper limit.   

Multi-task learning is a form of inductive transfer that can help improve a model by introducing an inductive bias, which causes a model to prefer some hypotheses over others\cite{Baxter2011}. In the instance of MTL, the auxiliary tasks supply inductive bias, causing the model to prefer hypotheses that explain more than one task\cite{Ruder2017}. The function of inductive bias is to choose for the model a hypothesis space large enough to contain a solution to the problem being learned, yet small enough to ensure reliable generalization from reasonably sized training sets.

As the difficulty of classification tasks increases, so does the amount of knowledge required for the model to introduce. This procedure improves the inductive abilities of the model and optimizes the hypothesis space, in addition to improving the quality of the generated texts. However, induction ability of the model has an upper limit. When the difficulty of the classification tasks surpasses the induction ability of the model, the induction result is incorrect, and the hypothesis space is not optimal. This explains why the quality of generated texts does not always improve with increasing difficulty of the classifications task.

\subsection{Category Accuracy of Generated Texts Impacted by Classification}

\begin{figure}[H]
	\centering
	\subfigure[composition of ICA-$K$C series datasets]{
		\includegraphics[width=2in]{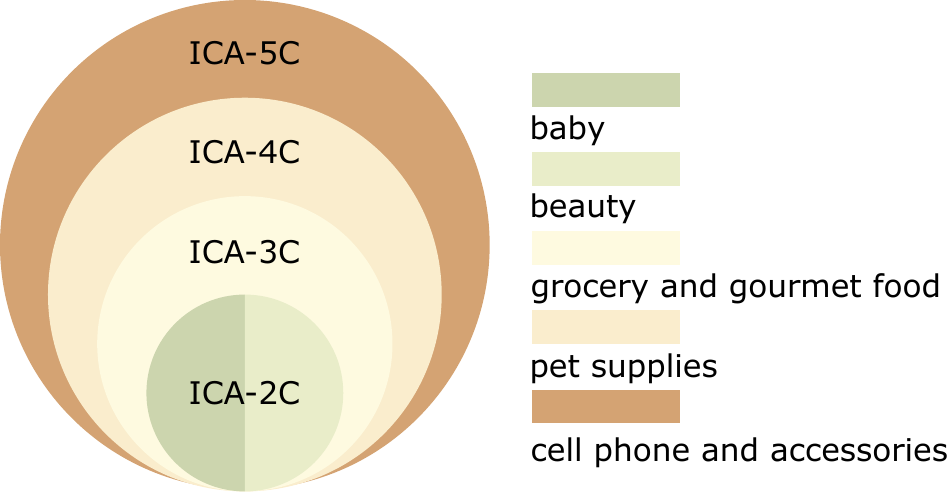}
		\label{fig.ar2_4}
	}
	\subfigure[category accuracy curves]{
		\includegraphics[width=1.5in]{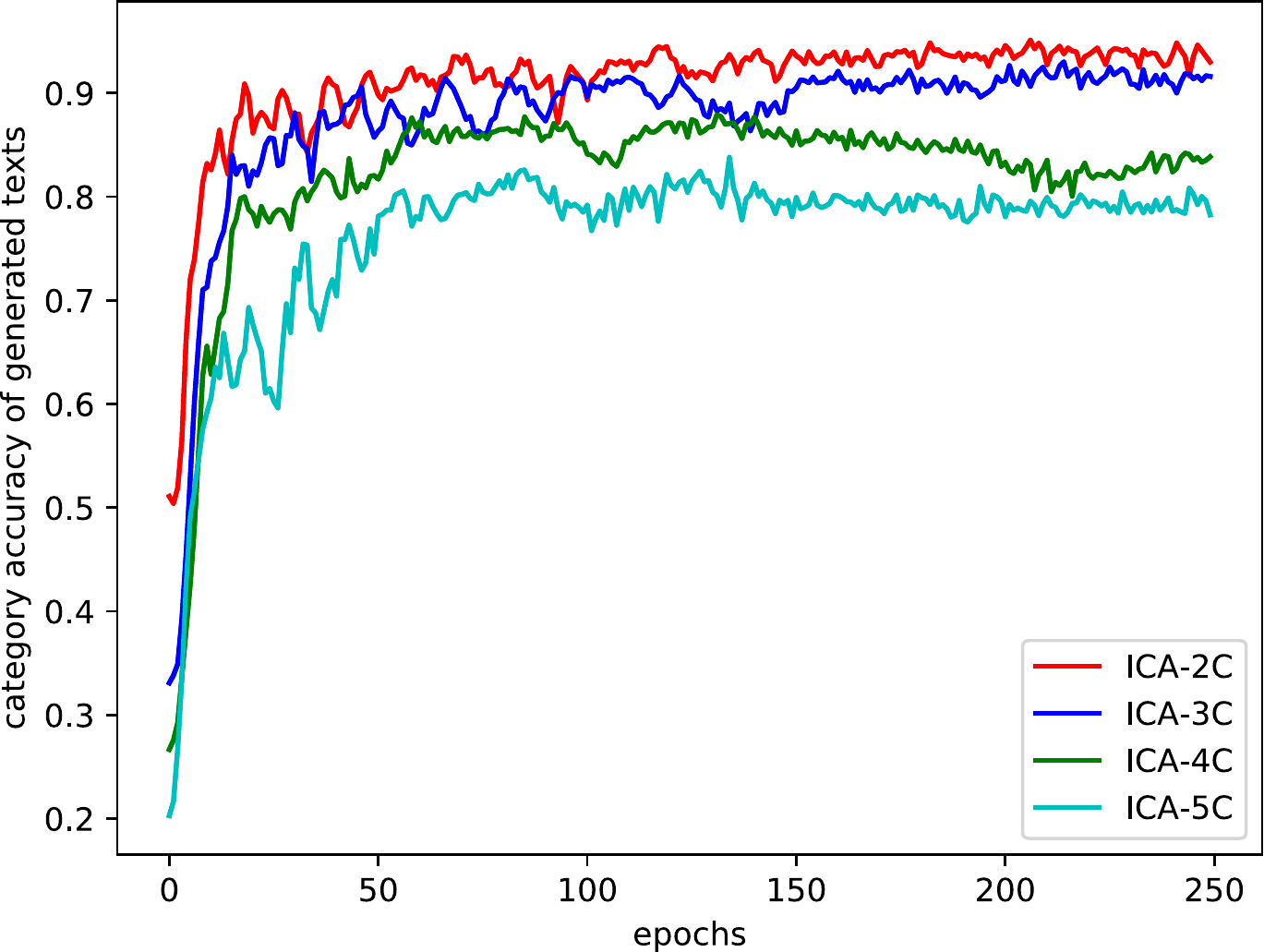}
		\label{fig.acc_categories}
	}
	\caption{The impact of classification difficulty on category accuracy. In (a) figure, each color corresponds to a product, and each circle corresponds to a dataset. The dataset represented by the inner circle is a subset of the dataset represented by the outer circle. The larger the circle, the more products there are in the dataset. In (b) figure, as classification difficulty increases, category accuracy decreases.}
	\label{fig.ar55}
\end{figure}

To illustrate the impact of classification on the category accuracy of generated texts, a series of datasets named the impact of classification on accuracy (ICA-$K$C) were built from Amazon reviews\cite{McAuley2015}, where $K$ is the number of categories. The samples and vocabulary of the ICA-$K$C series datasets grow in proportion to the number of categories. This design guarantees that the same category of products in multiple datasets share the same samples and vocabulary, to ensure that the category accuracy of texts generated by CatVRNN is only affected by the difficulty of classification. The details of the datasets are shown in Fig.\ref{fig.ar2_4}, where ICA-2C$\subseteq$ICA-3C$\subseteq$ICA-4C$\subseteq$ICA-5C and the sample size of a single product is 2000.

Similarly, the CatVRNN-adaptive model was adopted to train four datasets. As shown in Fig.\ref{fig.acc_categories}, the category accuracy of the generated texts decreases with increasing number of categories. As the number of epochs increased, the category accuracy curves tended to be stable; however, the category accuracy of the generated texts cannot be improved by increasing the number of training epochs.  

\section{Conclusion}
In this paper, CatVRNN is proposed to generate category texts. To guide the model to obtain category samples accurately, the idea of multi-task learning was applied to introduce classification tasks into the generation model. To further control the quality and diversity of generated texts, a function was designed to initialize the hidden state of CatVRNN. Extensive experiments demonstrated the efficacy of CatVRNN. Experimental results show that CatVRNN increases the diversity of generated text as much as possible while maintaining the fluency and category accuracy of the generated text. In addition, it has lower complexity and resource consumption than GANs. In future work, more complex and sophisticated VAE structures of VRNNs need to be explored to enhance the quality of the generated texts. The effective initialization of the hidden state of RNN is a worthy problem that needs to be studied.  

\section*{Acknowledgments}	
This work was supported by the Key Research and Development Program of Sichuan Province (Grant no.2020YFG0076).

\bibliographystyle{unsrt} 
\bibliography{catvrnn}

\end{document}